\documentclass[acmsmall,screen]{acmart}

\usepackage{hyperref}
\usepackage{listings}
\usepackage{subcaption}

\usepackage{color}

\definecolor{dkgreen}{rgb}{0,0.6,0}
\definecolor{gray}{rgb}{0.5,0.5,0.5}
\definecolor{mauve}{rgb}{0.58,0,0.82}

\AtBeginDocument{%
  \providecommand\BibTeX{{%
    \normalfont B\kern-0.5em{\scshape i\kern-0.25em b}\kern-0.8em\TeX}}}

\setcopyright{acmcopyright}
\copyrightyear{2024}
\acmYear{2024}
\acmDOI{10.1145/3649458}

\begin{document}

\title{Advancing Attribution-Based Neural Network Explainability through Relative Absolute Magnitude Layer-Wise Relevance Propagation and Multi-Component Evaluation}

\author{Davor Vukadin}
\email{davor.vukadin@fer.hr}
\orcid{0000-0003-3309-6718}
\affiliation{%
  \institution{University of Zagreb Faculty of Electrical Engineering and Computing}
  \streetaddress{3 Unska street}
  \city{Zagreb}
  \country{Croatia}
}

\author{Petar Afrić}
\email{petar.afric@fer.hr}
\orcid{0000-0001-9270-5988}
\affiliation{%
  \institution{University of Zagreb Faculty of Electrical Engineering and Computing}
  \streetaddress{3 Unska street}
  \city{Zagreb}
  \country{Croatia}
}

\author{Marin Šilić}
\email{marin.silic@fer.hr}
\orcid{0000-0002-4896-7689}
\affiliation{%
  \institution{University of Zagreb Faculty of Electrical Engineering and Computing}
  \streetaddress{3 Unska street}
  \city{Zagreb}
  \country{Croatia}
}

\author{Goran Delač}
\email{goran.delac@fer.hr}
\orcid{0000-0001-5315-8387}
\affiliation{%
  \institution{University of Zagreb Faculty of Electrical Engineering and Computing}
  \streetaddress{3 Unska street}
  \city{Zagreb}
  \country{Croatia}
}

\renewcommand{\shortauthors}{Vukadin et al.}

\begin{abstract}
  Recent advancement in deep-neural network performance led to the development of new state-of-the-art approaches in numerous areas. However, the black-box nature of neural networks often prohibits their use in areas where model explainability and model transparency are crucial. Over the years, researchers proposed many algorithms to aid neural network understanding and provide additional information to the human expert. One of the most popular methods being Layer-Wise Relevance Propagation (LRP). This method assigns local relevance based on the pixel-wise decomposition of nonlinear classifiers. With the rise of attribution method research,  there has emerged a pressing need to assess and evaluate their performance. Numerous metrics have been proposed, each assessing an individual property of attribution methods such as faithfulness, robustness or localization. Unfortunately, no single metric is deemed optimal for every case, and researchers often use several metrics to test the quality of the attribution maps. In this work, we address the shortcomings of the current LRP formulations and introduce a novel method for determining the relevance of input neurons through layer-wise relevance propagation. Furthermore, we apply this approach to the recently developed Vision Transformer architecture and evaluate its performance against existing methods on two image classification datasets, namely ImageNet and PascalVOC. Our results clearly demonstrate the advantage of our proposed method. Furthermore, we discuss the insufficiencies of current evaluation metrics for attribution-based explainability and propose a new evaluation metric that combines the notions of faithfulness, robustness and contrastiveness. We utilize this new metric to evaluate the performance of various attribution-based methods. Our code is available at: \url{https://github.com/davor10105/relative-absolute-magnitude-propagation}
\end{abstract}



\received{19 June 2023}

\acmJournal{TIST}
\maketitle

\section{Introduction}
Deep neural networks (DNNs) have become essential for processing diverse data types, enabled by recent advances in deep learning hardware. These developments allow training models with unprecedented scale in both parameters and training data size. Scaling these variables often leads to state-of-the-art models surpassing human-level performance in tasks like image classification \cite{Russakovsky2015}, reinforcement learning \cite{Silver2016} and natural language understanding \cite{deberta}. The presence of an extensive number of parameters and multiple non-linear layers contributes to the opaque nature of these models, rendering them challenging to interpret and understand. This fact encourages many researchers, especially in areas where model explainability is crucial, such as aviation, medicine or banking, to use smaller, often linear models that are transparent to their decision process.

In the realm of explainable artificial intelligence (XAI), two critical concepts are \textit{model interpretability} and \textit{interpretations}. Model interpretability assesses the intrinsic properties of a deep model, gauging the humans comprehensebility of its inference results. Interpretations involve algorithms explaining how deep models make decisions, such as highlighting discriminative features for a specific classification or identifying influential training examples. Given the paper's goal to unveil the inner workings of black-box models, the primary focus is on the concept of \textit{interpretations}.

Various interpretation tools aim to explain DNN decision-making, but no method suits every task and model, making this area of research ongoing. Different principles create interpretability methods, such as highlighting input features on which the deep model mainly relies for a particular inference, either by using gradients \cite{https://doi.org/10.48550/arxiv.1311.2901} \cite{https://doi.org/10.48550/arxiv.1706.03825} \cite{ancona2018towards}\cite{https://doi.org/10.48550/arxiv.1703.01365} \cite{Selvaraju_2019}, perturbations \cite{https://doi.org/10.48550/arxiv.1806.07421} or proxy explainable models \cite{https://doi.org/10.48550/arxiv.1602.04938}; visualizing intermediate features \cite{https://doi.org/10.48550/arxiv.1311.2901} or visualizing counterfactual examples \cite{wachter2017counterfactual}; analyzing training data to assess each data point's contribution to a particular inference result \cite{https://doi.org/10.48550/arxiv.1703.04730}. This paper's focus is on highlighting the input features that the model relied on most during inference, by propagating and tracking layer activations through the deep network.

Subjective evaluation of attribution maps can be conducted by humans, but a quantitative measure is crucial for objective comparison and ranking of different attribution methods. Our focus is on evaluating explanation quality for specific examples and neural networks, acknowledging the complexity of this task. Explanation quality depends not only on the attribution method but also on classifier performance influenced by factors like network architecture and training data.

Attributions should reflect the model's perspective rather than strictly adhering to human intuition. A segmentation map of an object may not qualify as an attribution map, potentially overlooking crucial evidence. Evaluation methods categorize different aspects like faithfulness, robustness, and localization. However, there's no established approach in the literature to effectively combine scores across classes for determining the best attribution method.

Our work addresses current evaluation method limitations, proposing a novel approach that integrates faithfulness, robustness, and localization into a single score. This comprehensive assessment aims to inform researchers and practitioners about the performance of attribution methods, facilitating informed decisions in their selection and application.

To summarize, the main contributions of this work are:
\begin{itemize}
 \item The development of a novel Layer-Wise Propagation rule, referred to as Relative \textbf{Abs}olute Magnitude \textbf{L}ayer-Wise \textbf{R}elevance \textbf{P}ropagation (absLRP). This rule effectively addresses the issue of incorrect relative attribution between neurons within the same layer that exhibit varying absolute magnitude activations. We apply this rule to three different architectures, including the very recent Vision Transformer, and demonstrate clear advantages over existing work.
 \item The proposal of a new evaluation method, \textbf{G}lobal \textbf{A}ttribution \textbf{E}valuation (GAE), which offers a novel perspective on evaluating faithfulness and robustness of an attribution method by utilizing gradient-based masking, while combining those results with a localization method to achieve a comprehensive evaluation of explanation quality in a single score. To validate our approach, we compare absLRP against various state-of-the-art and commonly used attribution methods using three different architectures: VGG, ResNet50, and ViT-Base. We conduct experiments on two widely used image classification datasets, namely ImageNet and PascalVOC. The results clearly demonstrate the superiority of our method and provide valuable insights into the strengths and weaknesses of each attribution method.
 \end{itemize}

\section{Related work}\label{rel_sec}
In the following section, we offer a comprehensive overview of the existing literature on layer-wise relevance propagation, transformer explainability, and attribution-based evaluation metrics.
\subsection{Layer-Wise Relevance Propagation}
Bach et al. \cite{Bach_2015} presented an algorithm for determining the relevance of a particular input neuron to the output of a non-linear network called \textit{layer-wise relevance propagation} (LRP). LRP assumes that the classifier can be decomposed into several layers of computations, which can be parts of the feature extraction phase or parts of a classification algorithm that runs on the computed features.

Given an input $x$ and a neural network $f$ the aim of LRP is to assign each input position $p$ (for example, in the case of images, to each pixel) a relevance score $R_p^0$, where the superscript $0$ indicates the first, input layer. Assuming the knowledge of the relevance map of the last layer $R_p^l$, the goal of a particular LRP formulation is to describe how to disperse output neuron $j$ relevance to each of the input neurons $i$ in the layer before - $R_{i \leftarrow j}^{(l-1, l)}$, such that the following equation holds:
\begin{equation}
    R_i^{(l-1)} = \sum_{j \in (l)} R_{i \leftarrow j}^{l}
\end{equation}
For a particular layer of a neural network with an activation function $g$ and input $x$, the activation of an output neuron in the following layer is defined as:
\begin{equation}
    a_j = g(\sum_i x_iw_{ij})
\end{equation}
The term $x_0$ is set to 1 so that $w_{0j}$ represents the bias of the output neuron $j$.
Bach et al. \cite{Bach_2015} proposed several formulas for computing $R^{l-1}$ from relevance scores of the following layer $R^{l}$:

\begin{equation}
    \epsilon\text{-rule} \text{(LRP-}\epsilon\text{):} \hspace{1cm}
    R_{i}^{l-1} = \sum_j \frac{x_i w_{ij}}{\sum_k x_k w_{kj} + \epsilon} R_j^l
\end{equation}

This rule redistributes the relevance of the following layer to the layer before based on the proportion of the contributions of each input neuron to the activation of the output neuron.

\begin{equation}
    \alpha \beta\text{-rule (LRP-}\alpha \beta\text{):} \hspace{1cm}
    R_{i}^{l-1} = \sum_j (\alpha \frac{(x_i w_{ij})^+}{\sum_k (x_k w_{kj})^+} - \beta \frac{(x_i w_{ij})^-}{\sum_k (x_k w_{kj})^-}) R_j^l
    \label{eq:alphabeta}
\end{equation}
LRP-$\alpha \beta$ treats positive and negative activations separately, offering new terms, hyperparameters of this method, $\alpha$ and $\beta$ to specify the relative importance between the two categories. Bach et al. \cite{Bach_2015} have found that values of $\alpha = 2$ and $\beta = 1$ produce sharp relevance maps. 

However, the authors of Relative Attributing Propagation (RAP) \cite{nam2019relative} highlight an issue with commonly used LRP rules when propagating relevance towards the input layer. If a neuron receives conflicting large relevance values (e.g., one large positive and one large negative) from its subsequent layer, these values tend to cancel out, despite their significant impact on the model's final output. This leads to sparse and challenging-to-interpret attribution maps, with dominance in only a few locations where one contribution prevails. RAP tackles this problem by combining positive and negative attributions based on the relative magnitude within the absolute value of each contribution. Essentially, RAP performs two LRP-$\alpha_1\beta_0$ passes, one for the positive part and one for the negative part of the subsequent layer, weighs them by their relative size within the absolute sum, and then adds them together. To address potential over-allocation of relevance, RAP subtracts the mean difference between the previous and current layer's relevance sum from all non-zero assigned relevances after each layer. The formal definition of RAP is:

\begin{equation}
\begin{gathered}
    R_{i \in \mathcal{P}, \mathcal{N}}^{l-1} = \sum_j (\alpha \frac{(x_i w_{ij})^+}{\sum_k (x_k w_{kj})^+} + \beta \frac{(x_i w_{ij})^-}{\sum_k (x_k w_{kj})^-}) R_j^l
\\
\alpha = \frac{\sum_i (x_i w_{ij})^+}{\sum_i ((x_i w_{ij})^+ + |(x_i w_{ij})^-|)}
\hspace{1cm}
\beta = \frac{\sum_i (x_i w_{ij})^-}{\sum_i ((x_i w_{ij})^+ + |(x_i w_{ij})^-|)} \\
R_{i}^{l-1} = R_{i \in \mathcal{P}, \mathcal{N}}^{l-1} - \Psi^{l-1}_i
\end{gathered}
\end{equation}

where $\Psi^{l-1}_i$ is the mean value of non-zero-neurons in the layer $l-1$.

\begin{figure}
     \centering
     \includegraphics[width=0.9\columnwidth]{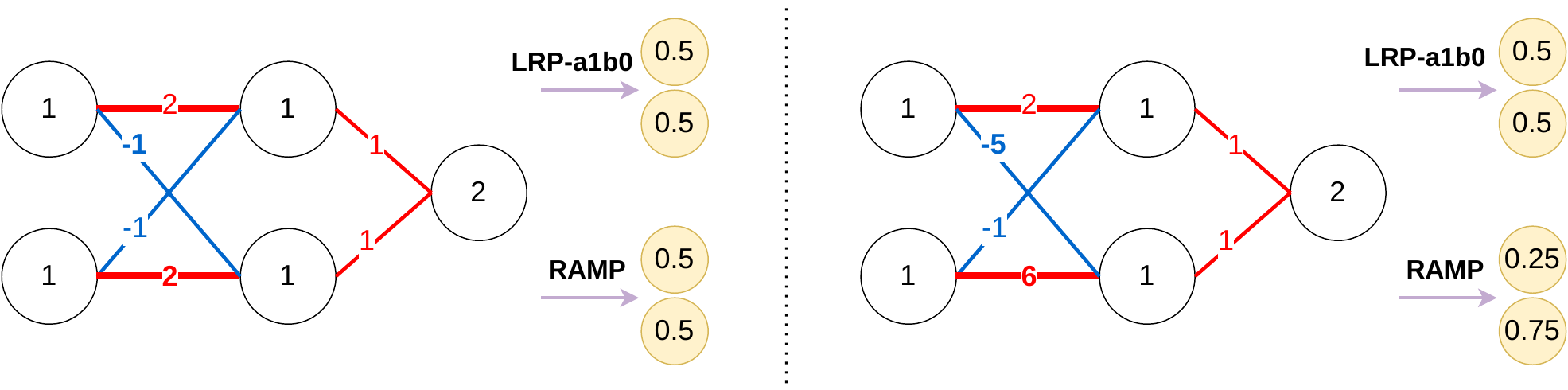}
     \Description{Two neural networks side-by-side with two feature inputs, two hidden neurons and one output neuron. For the network on the left, weights going from input to hidden neurons are 2 and -1; and -1 and 2 respectively. Both weights from hidden neurons to the output are 1. For the right network, input weights are 2 and -5; and -1 and 6. Weights from hidden neurons to the output are again both 1.}
     \caption{Toy examples illustrating key differences between LRP-$\alpha_1\beta_0$ and our method. 
White circles represent inputs and layer activations, with weights on connecting lines, while final attribution values are shown in yellow circles. In the left example, equal relative absolute magnitudes result in similar attributions for LRP-$\alpha_1\beta_0$ and our method. In the right example, vastly different activation magnitudes lead our method to consider the larger absolute activation of the bottom hidden neuron in the attribution map.}
     \label{fig:intline_vs_alpha1beta0}
\end{figure}

There are, however, two main issues with RAP. The first concerns LRP-$\alpha_1\beta_0$. Examining the propagation equation for the LRP-$\alpha\beta$ (Equation \ref{eq:alphabeta}), we observe that for the positive ($\alpha$) part of the attribution, the denominator contains only the positive part of the next layer's value, and conversely the negative ($\beta$) part only contains the sum of layer's negative values. This, we hypothesize, leads to poor relative assignment of attributions when there is a large amount of both positive and negative activations in the layer. Illustrated in Figure \ref{fig:intline_vs_alpha1beta0}, both toy examples share the same architecture, input, hidden and output values. However, for the right network, the second hidden neuron has larger magnitude weights connecting to it ($[-5, 6]$). Using the LRP-$\alpha_1\beta_0$ rule and propagating backwards towards the input for both networks, we obtain the same attribution map of $[0.5, 0.5]$, when normalized such that the absolute sum of the attribution scores is equal to 1. The concept of uniform attribution across input neurons contradicts the underlying dynamics within the network. This is evident when examining the absolute magnitudes of activations in the second hidden neuron, significantly larger at 11 compared to the first neuron's magnitude of 3. In such cases, it's reasonable to expect the second input neuron to carry more positive relevance, with its positive value of 6 being three times greater than the first neuron's positive relevance of 2. This gap between the attribution map and the actual layer dynamics increases with the difference between absolute activation magnitudes and the neuron's output value. Therefore, a more effective layerwise propagation approach should account for the absolute magnitudes of activations within a neuron.

The second issue is found in the over-allocation relevance reassignment. As the first issue results in an over-allocation in relevance for the current layer, the decision to evenly subtract the over-allocated difference is arbitrary, lacking justification from the authors. In our work, we rely on contrastive relevance propagation to discern important parts of the image.

\subsection{Transformer Explainability}
Transformers have become a staple in machine learning research and state-of-the-art approaches in almost all fields, such as image \cite{vision_transformer} and natural language processing \cite{bert}, image generation \cite{dalle} and reinforcement learning \cite{rl_transformer}. Their widespread use necessitates the development of explainability tools which can aid in model debugging and verifying fairness and unbiasedness of such approaches.

The main building blocks of Transformers are the self-attention layers, mapping a query and key-value pairs to an output. Several works exploited these attention scores as relevancy scores \cite{vaswani} \cite{https://doi.org/10.48550/arxiv.2005.12872} \cite{https://doi.org/10.48550/arxiv.1502.03044}, but only using the final attention layer to compute these scores. Another option is combining multiple layers to produce the final relevance output. Simple averaging of per-layer relevance score would produce a blurred relevance signal and would not account for the different roles of each layer \cite{beyond_transformer}. Attention Rollout \cite{rollout} recursively computes the token attentions in each layer of a given model given the embedding attentions as input. This method can be defined as:
\begin{equation}
\begin{gathered}
    \text{Rollout} = \hat{\textbf{A}}^{1} \times \hat{\textbf{A}}^{2} \times ... \times \hat{\textbf{A}}^{B}
    \hspace{1cm}
    \hat{\textbf{A}}^{b} = \textbf{I} + \mathbb{E}_h \textbf{A}^{b}
\end{gathered}
\end{equation}
Where $\textbf{A}^{l} \in \mathbb{R}^{h \times s \times d_h}$ denotes the attention map in block $b$. $\mathbb{E}_h$ is the average value across the head dimension. Chefer et al. \cite{beyond_transformer} augment this approach by using LRP-$\epsilon$ ($\epsilon = 10^{-9}$) to calculate the relevance of self-attention tensor at each layer $\textbf{R}^{\textbf{A}^b}$ multiplied by the gradient of the self-attention tensor $\nabla \textbf{A}^{b}$ with respect to a target class. Furthermore, only positive relevances are taken into account per each layer. This method is defined as:
\begin{equation}
\begin{gathered}
    \text{TIBAV} = \hat{\textbf{A}}^{1} \times \hat{\textbf{A}}^{2} \times ... \times \hat{\textbf{A}}^{B}
    \hspace{1cm}
    \hat{\textbf{A}}^{b} = \textbf{I} + \mathbb{E}_h(\nabla \textbf{A}^{b} \odot \textbf{R}^{\textbf{A}^b})^{+}
\end{gathered}
\end{equation}

In all the aforementioned methods, the sole use of attention matrices as the means of calculating final attribution maps results in their reduced fidelity, as the resolution of their attribution maps is dependent on the number of patches used by the model. For example, a standard Vision Transformer with patch size $16 \times 16$ processing a $224 \times 224$ image would produce $14 \times 14$ patches, which are then simply upscaled to the original resolution. In the case of Vision Transformers with fewer, larger patches designed for speed and efficiency, patch-wise attribution methods remain futile.

To the best of our knowledge, only two studies report the usage of LRP-only methods in Transformer model explainability. Chefer et al. \cite{beyond_transformer} use LRP-$\alpha_{1}\beta_{0}$ throughout the model, but report poor results and use this method as a baseline only. Ali et al. \cite{xai_bert} embed the Gradient $\times$ Input method \cite{gradxinput} into the LRP framework by identifying LRP conservation rule breaks that this method introduces in self-attention and layer normalization layers. The self-attention conservation issue is solved by viewing the softmax self-attention scores as constants and not propagating the attributions through them, similarly to the gating value used in the LRP for LSTMs formulation \cite{lstm_lrp}. Additionally, the results are not reported in the domain of computer vision, but only in NLP and graph processing.
 
 We resolve all the aforementioned issues by applying our new LRP rule to all layers of the Vision Transformer and demonstrate a clear advantage over existing methods.

\subsection{Attribution-based Evaluation Metrics} \label{evaluation_metrics}

The evaluation of attribution-based explanation methods can be categorized into several classes, each assessing different properties of these methods. The following classes are notable.

\textbf{Faithfulness} - consists of \textit{importance correlation} - the magnitude of attribution weights should reflect the importance of input components and \textit{polarity consistency} - the sign of attribution weights should correctly indicate the polarity of input impact, in other words, the contribution or suppression effects to the model's prediction, as defined by Liu et al. \cite{liu2022rethinking}. Metrics in this category are based on observing the change in the model's output when permuting the input based on a given attribution map. Several metrics aim to compute the correlation between the model's original probability output and the perturbed one \cite{https://doi.org/10.48550/arxiv.2005.00631} \cite{NEURIPS2018_3e9f0fc9} \cite{https://doi.org/10.48550/arxiv.2007.07584} \cite{Montavon_2018} \cite{ancona2018towards} \cite{https://doi.org/10.48550/arxiv.2003.08747} \cite{https://doi.org/10.48550/arxiv.1901.09392} or the correlation between probability drops and attribution scores on various points \cite{NEURIPS2018_3e9f0fc9} \cite{https://doi.org/10.48550/arxiv.2005.00631}, while others observe the change in model performance after perturbing the input \cite{https://doi.org/10.48550/arxiv.1909.03012} \cite{2202.00449} \cite{lee_evaluation_metric_haas}.

\textbf{Robustness} - measures the stability of attributions when subject to slight perturbations of the input, assuming that model output approximately stayed the same. These can be further divided in metrics that look for similar examples that should produce similar explanations \cite{https://doi.org/10.48550/arxiv.1901.09392}, capture the strongest variation in the attribution map when perturbing the input \cite{Montavon_2018} or measures the probability that the inputs with the same attributions have the same prediction label \cite{https://doi.org/10.48550/arxiv.2202.00734}.

\textbf{Localization} - quantify the alignment between the highest scores of the generated attribution map and the target object outlined by the ground-truth mask \cite{https://doi.org/10.48550/arxiv.1608.00507} \cite{https://doi.org/10.48550/arxiv.1910.09840} \cite{https://doi.org/10.48550/arxiv.2003.07258}.

We outline three most relevant approaches for our evaluation metric and address their drawbacks.

\textbf{Approach 1.} Samek et al. \cite{samek2016evaluating} propose a metric for evaluating an attribution-based explanation method as a generalization of the approach presented by Bach et al. \cite{Bach_2015}, where the authors use attributions generated by different methods to guide region perturbation of the input image. Concretely, for a model $f$, in each of the $L$ steps indexed by $k$, the first $k$ $9 \times 9$ regions of the input image sorted by attribution scores undergo perturbation by replacing them with randomly sampled values from a uniform distribution. At each step, a difference in the output neuron's value is observed between the original value and the value gotten by passing the perturbed image at step $k$: $x_{MoRF}^k$, where MoRF is an abbreviation of the used process - \textit{most relevant first}. Those differences are then added together for every step and averaged over the entire dataset. An ordering of regions such that the most sensitive regions are ranked first would imply a steep decrease of the output neuron's value, this resulting in a larger \textit{area over the MoRF perturbation curve} (AOPC):
\begin{equation}
    \textrm{AOPC} = \frac{1}{L + 1} \frac{1}{N} \sum_n^N \sum_k^L (f(x_{n_{MoRF}}^0) - f(x_{n_{MoRF}}^k))
\end{equation}

As a further extension of this metric, the authors propose another metric, \textit{area between perturbation curves} (ABPC). This metric takes in account the gap between the output neuron scores for an image modified by perturbing the most relevant first (MoRF), and the least relevant first (LeRF):
\begin{equation}
    \textrm{ABPC} = \frac{1}{L + 1} \frac{1}{N} \sum_n^N \sum_k^L (f(x_{n_{LeRF}}^k) - f(x_{n_{MoRF}}^k))
\end{equation}
In the case of LeRF, the information in the image should be very stable and close to the original value for small value of $k$, and only drops quickly as $k$ approaches $L$. Thus, this metric gauges how good an attribution method is at selecting both the most relevant parts of the image, so that the classification score drops the most, while also providing information about the least relevant parts of the image, whereby perturbing them, the classification score changes the least. The authors perturb the first 100 regions of the image, resulting in 15.7\% of the image being exchanged.

Recently, Rong et al. \cite{2202.00449} expanded upon this approach. They followed the same principle of masking input features, which are either highly relevant (MoRF) or lowly relevant (LeRF), based on the initial attribution map. However, they introduced a new perturbation method by utilizing a linear combination of neighboring pixels. Their approach enhances the ranking consistency of the attribution methods when comparing the MoRF and LeRF masking methods.

Samek et al.'s metric, similar to other faithfulness metrics, relies on meaningful perturbations. The core idea is that alterations to the input, guided by the generated attribution maps, should significantly impact the baseline prediction if the attribution is faithful to the target model. However, as highlighted by Ju et al. \cite{ju2022logic}, this evaluation method can be viewed as an attribution method calculating a score for the k\% modified features. Thus, evaluation metrics that fall into this category actually calculate the similarity between two attribution methods instead of measuring faithfulness. Additionally, the magnitude of attributing scores of each input feature can vary greatly while still producing the same ordering. This can completely change the perception of the input feature's relative importance, while maintaining the exact same score on these types of evaluation methods. And finally, attribution methods need not explain how the input should be modified to produce a certain effect on the network's output, but instead need to correctly attribute the input feature's importance for the \textbf{current} input example, a notion that seems to be forgotten in previous work.

\textbf{Approach 2.} Alvarez et al. \cite{alvarez2018robustness} investigate the robustness of attribution methods, emphasizing that similar inputs should yield similar explanations. They propose a performance metric to quantify this idea, considering variations in the explanation of a prediction concerning changes in the input leading to that prediction. Robustness, in this context, relates to the consistency of attributions when the input is slightly perturbed, and the output changes only marginally or remains unchanged. The authors leverage the concept of Lipschitz continuity, viewed on a local scale through the solution of a specific optimization problem for each example of interest $x_{orig}$:
\begin{equation}
\begin{gathered}
    \hat{L}(x_{orig}) = \underset{x_{pert} \in \aleph_{\epsilon}(x_{orig})}{argmax} \frac{||f(x_{orig}) - f(x_{pert})||_2}{||x_{orig} - x_{pert}||_2} \\
    \aleph_{\epsilon}(x_{orig}) = \{ ||x_{orig} - x_{pert}||_2 \le \epsilon ;  x_{pert} \in \mathbb{R}^n \}
\end{gathered}
\end{equation}
Optimized value $\hat{L}(x_i)$ is viewed as the measure of robustness.

There are, however, several issues associated with this metric. Attribution methods inherently differentiate between more and less important input features. By perturbing all features, even to a small extent as proposed in the detailed approach, there is a risk of disrupting highly influential input features. Such perturbations could potentially lead to a complete alteration of the model's reasoning process without affecting the prediction outcome. This concern was demonstrated in the study conducted by Ju et al. \cite{ju2022logic}, where they sought adversarial examples that minimized similarity in the selected subset, rather than focusing solely on model predictions. Additionally, it is important to note that an attribution method that achieves a perfect score according to this evaluation metric would be a constant. This further undermines the reliability and validity of such metrics.

\textbf{Approach 3.} Arias-Duart et al. \cite{arias2022focus} define a novel metric called the Focus, which uses compositions of images from the dataset, called mosaics, to measure the amount of relevance allocated by the relevance method to the correct (target) regions of the mosaic. Each mosaic $m$ is constructed by selecting a unique class which the attribution method is expected to explain. Then, two images that belong to that class are selected and deemed as positives ($p_1, p_2$), and two images that do not belong to the selected class are randomly sampled and deemed as negatives ($n_1, n_2$). These four images are then combined into one image by combining them as a two-by-two, non-overlapping grid, with the positions of each image being chosen randomly.  Finally, the Focus metric is defined as:

\begin{equation}
    F(m) = \frac{R(p_1) + R(p_2)}{R(m)}
\end{equation}

$R$ is the sum of positive target class relevance in the respective quadrant, or the whole mosaic.

The Focus metric by Arias-Duart et al. \cite{arias2022focus} is a part of the localization approach in evaluating attribution maps. The idea is that a reliable attribution method should be able to concentrate most of its relevance on the two positive images belonging to the target class. The issue with this method is similar to our first detailed approach. Multiple very different attributions can result in the same exact Focus score. For example, an attribution map that correctly identifies the two squares with image positives and attributes the center pixel of each of those image positives as 1 would get the same, maximum Focus score as an attribution method that attributed the score 1 to the entire quadrants of the positive parts in the mosaic. This indicates the need of evaluating the attribution quality of a particular quadrant first, and then combining this score with this localization metric.
\\ \\
Challenges in attribution method evaluation metrics have led to a lack of consensus among researchers on the optimal metric. Consequently, researchers frequently use multiple metrics to assess different aspects of their methods. Effectively combining these diverse scores into a comprehensive ranking remains unclear, limiting widespread adoption and requiring users to experiment and subjectively choose the most suitable method for their model or specific problem.

\section{Method} \label{formulation_sec}
This section delves into a thorough description of our novel attribution approach, elucidating its intricacies and highlighting its applicability to complex architectures, including ResNets and Vision Transformers. Additionally, we introduce a novel evaluation metric for attribution methods, which effectively assesses their local and contrastive properties.
\subsection{Relative Absolute Magnitude Layer-Wise Relevance Propagation}
Motivated by the aforementioned issues with current LRP propagation rules, we propose a new rule called Relative \textbf{Abs}olute Magnitude \textbf{L}ayer-Wise \textbf{R}elevance \textbf{P}ropagation (absLRP). It solves the issue of conflicting relevance found in most LRP rules by observing only the positive parts of neuron activation, similarly to LRP-$\alpha_1\beta_0$. However, to address incorrect relative attribution due to varying magnitudes within neurons of the same layer, we use the absolute \textbf{final} output of each neuron as the normalizing factor. To produce contrastive attribution maps, we utilize the idea from Gu et al. \cite{gu2019understanding} and we set the last layer's starting attribution as $1$ for the target class, and $-\frac{1}{N}$ for the other classes, $N$ being the total number of classes. Applying this new rule yields sparse and contrastive attribution maps with no noise. The new LRP rule is defined as:

\begin{equation}
    \begin{split}
        R_{i}^{l-1} = \sum_j \frac{(x_i w_{ij})^+}{|\sum_k x_k w_{kj}| + \epsilon} R_j^l
    \end{split}
\end{equation}

Looking back to the example in Figure \ref{fig:intline_vs_alpha1beta0}, we already discussed how LRP-$\alpha_1\beta_0$ assigns the same relevance score to each of the input neurons, regardless of the relative absolute magnitudes of activations in the first and the second hidden neuron. By applying our rule on the left network, we obtain the same attribution scores as the LRP-$\alpha_1\beta_0$ method, since both of the branches have equal absolute magnitudes of 3. However, when applying our method to the right network, the resulting attributions become $[0.25, 0.75]$, a 50\% difference attribution difference for each of the input neurons compared to the LRP-$\alpha_1\beta_0$'s attribution, since the second branch now has a much larger absolute magnitude of 11 which leads to its greater impact on the attribution map. The magnitude difference between the two branches amplifies this effect. If we make the difference even larger, for example, by setting the second hidden neuron's weights as $[-17, 18]$, while LRP-$\alpha_1\beta_0$ keeps the same attributions, our method now assigns 9 times more relevance to the second input neuron, since its absolute positive influence in the network is 18, compared to the first neuron's 2.

This rule can be efficiently implemented via automatic differentiation, available in most neural network libraries. PyTorch-like pseudocode implementing our rule is presented in Algorithm \ref{alg:method_implementation}.

\begin{lstlisting}[caption={Pseudocode for calculating attribution maps based on our proposed rule},label=alg:method_implementation]
def backprop_relevance(prev_rel, input, module):
    """
    prev_rel : tensor
        A tensor representing next layer's relevance
    input : tensor
        Input tensor for this layer
    module : nn.Module
        This layer's implementation along with its parameters (if any)
    """

    # Calculate output
    h = module(input)

    # Calculate absolute output
    input_abs = input.abs()
    module_abs = abs_module_params(module)
    ha = module_abs(input_abs)

    # Backprop to obtain this layer's relevance
    rel_scaling = prev_rel / (h.abs() + 1e-9)
    rel = autograd.grad(ha + h, x, rel_scaling) * x

    return rel

\end{lstlisting}

Since our method relies on LRP, the process of adapting our method to more complex models primarily entails incorporating this rule into specific layers of the architecture. Notably, this can be accomplished seamlessly by integrating the rule into \textbf{residual connections} and \textbf{batch normalization} within ResNets or by applying it to \textbf{self-attention blocks} and \textbf{layer normalization} in Transformers. Implementing our method in these cases is straightforward.

\textbf{Residual Connection} - The LRP rule must be specified for two key stages in the computation graph. Firstly, it is essential to define the rule for the initial split of the residual and the standard components of the residual connection. Secondly, the rule needs to be established for the recombination stage at the end of the residual block in ResNets or the attention/feed-forward block in Transformers. To begin with, the recombination, the rule is simply defined as two passes through the new LRP rule. The first pass pertains to the standard part of the connection, while the second pass relates to the residual part of the connection. Following this, the standard part is backpropagated through the network in a conventional manner until it reaches the initial residual split point. At this juncture, we employ the remaining part of the residual rule, which involves summing up the two relevances provided by the respective components of the residual connection.

\textbf{Batch Normalization} - In this scenario, we treat the learned mean and standard deviation scaling, as well as the point-wise weight and bias, as two separate layers of operations. As a result, we apply our rule in a layer-wise manner to generate the attribution of this layer's input.

\textbf{Layer Normalization} - In this context, the mean and variance of the input utilized in Layer Normalization are detached from the computation graph and treated as constants. Subsequently, our LRP rule is applied to compute the relevance of the input, similarly to the batch normalization.

\textbf{Self-Attention Block} - This block comprises multiple sub-operations, each necessitating a backward LRP pass: self-attention multiplication between queries and keys, softmax operation for self-attention scores, and multiplication of these scores with values to yield the final output of the block. Similar to the residual connection approach, we perform separate backward passes for each input of the operation. Specifically, one pass for queries and another for keys in self-attention multiplication, and separate passes for the attention score matrix and values in the multiplication operation that produces the final output. For the softmax LRP pass, our established rule is once again employed. However, in this case, the denominator of softmax, containing the sum of exponentiated inputs, is detached from the computation graph, similar to how the mean and variance values were detached in the Layer Normalization layer. To combine the attributions from queries, keys and values, as was the case in the residual connection, we sum up their attributions.

\subsection{Global Attribution Evaluation}
To address existing issues in attribution map evaluation, we propose a comprehensive metric for assessing the quality of a given attribution method $m_A$ by observing several factors at once and combining them into a single score. These factors are \textbf{Local consistency and Contrastiveness}.

To begin the process, we randomly sample four images from the dataset. Among these four images, we select one at random and designate it as the positive image. Subsequently, we calculate the Local consistency score for the positive image.

\textbf{Local consistency} - this is a factor unexplored in the existing literature, as the majority of research has primarily concentrated on two desirable aspects of attribution methods: faithfulness and robustness. In our work, however, we measure both of those factors in a novel way.

In contrast to previous studies that either relied on the ranking of input features provided by the initial attribution map or used a random subset of features for perturbation, our approach employs a gradient-based method at each step to identify areas that will result in the most significant or least significant change in the output score. Specifically, at each step (denoted as $t$), we select the top $k$ percent of input features based on the gradient-based approach.

In our experiments, we utilize zero masking as the perturbation method, employing the absolute value of the product between the input value and the gradient of our masking loss with respect to the input. The masking loss is simply defined as the absolute value of the target output. This strategy leverages the insight that larger gradient values with respect to the input correspond to areas of the image that have the greatest impact on the model's output. Multiplying these values with the input provides an estimate of the impact of zeroing out those positions on the model's output. This alternative approach addresses a limitation in existing research on evaluation metrics. In previous studies, the attribution map, which is expected to explain relevant features for a given input example, is also employed for an unrelated purpose – identifying input features that cause the largest change in the model's output. Although these notions may partially overlap for some input features, they are distinct and should be treated as such. By using our proposed method, we can alleviate this issue and ensure a clear distinction between the two objectives.

We perform $T$ steps of perturbation by selecting the features with the most significant effect first - Most Relevant First approach (MoRF). Conversely, we also perform an equal number of steps, but this time selecting features with the lowest impact first - Least Relevant First approach (LeRF).

As a consequence, two sequences of model output values are generated. The first sequence corresponds to the step-wise perturbation of the most impactful features, and it is anticipated that the output value $o_{t}$ at each step will exhibit a rapid decrease from the initial output value $o_{init}$. Conversely, the second sequence represents the step-wise perturbation of the least impactful features, where the output value is expected to decrease at a slower pace.

In the subsequent section, we compute the attribution maps for each perturbed input at each step, considering both MoRF and LeRF modes of masking. We then assess the similarity between these maps and the initial one, quantified by the following definition:

\begin{equation}
    \text{sim}_t^{Mo/Le} = 1 - \frac{|| A_{init} - A_{t}^{Mo/Le} ||_1}{||A_{init}||_1 + ||A_{t}^{Mo/Le}||_1}
\end{equation}

where $||A||_1$ denotes the $L^1$-norm of the flattened tensor.

Again, this results in two sequences that describe how similar the attribution maps were at each step of the perturbation, when perturbing the largest impact features, and when perturbing the lowest impact features. Similarly to the output, it is expected that the similarity between the initial attribution map and current step's map should drop quickly when perturbing the most impactful features, and conversely, change slowly when doing the opposite. Moreover, the difference between these two sequences - $d^A$ should correlate with the difference between the two output value sequences - $d^o$. Given that we are masking crucial evidence associated with a specific class, it is expected that the attribution map will gradually deviate from its initial state as this evidence is removed, resulting in a decrease in the output value. Conversely, when perturbing the least impactful features and retaining only the most important evidence in the input, the attribution map should remain relatively stable, with the output value remaining close to its initial value.

To compute the correlation between these two curves, we normalize all the output values by the initial output value. This normalization step is crucial to ensure that the metric is normalized per example, as different images can have varying initial outputs. Normalizing the values helps prevent bias towards classes with higher or lower final outputs. Once the normalization is performed, we proceed to calculate the robustness component of the local consistency score:


\begin{equation}
\begin{gathered}
    \textbf{LC}_R(m_A) = 1 - \frac{2 || d^o - d^A ||_1}{|| d^o ||_1 + || d^A ||_1} \hspace{1cm}
    d^o_t = o_t^{Le} - o_t^{Mo} \hspace{1cm}
    d^A_t = sim_t^{Le} - sim_t^{Mo}
\end{gathered}
\end{equation}

\begin{figure}
     \centering
     \begin{subfigure}[b]{0.3\columnwidth}
         \centering
         \includegraphics[width=\textwidth]{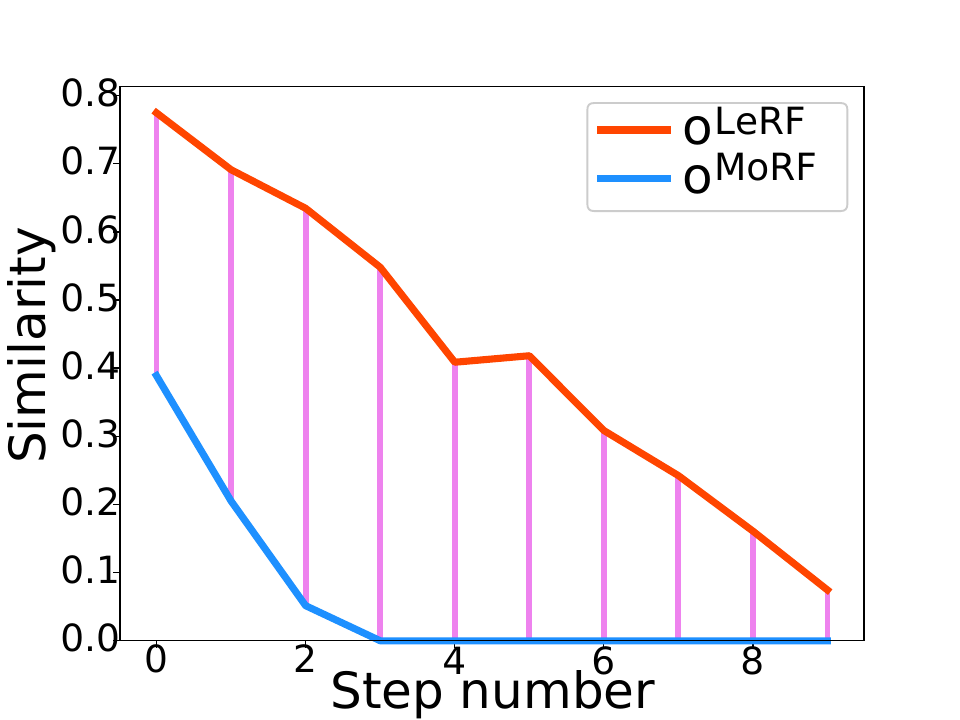}
         \Description{A graph with two line plots representing the output MoRF and LeRF change curves. Both curves are dropping down and to the right as more input neurons are masked. However, the MoRF curve drops significantly faster than the LeRF curve.}
         \caption{}
         \label{fig:morf_lerf_out}
     \end{subfigure}
     \begin{subfigure}[b]{0.3\columnwidth}
         \centering
         \includegraphics[width=\textwidth]{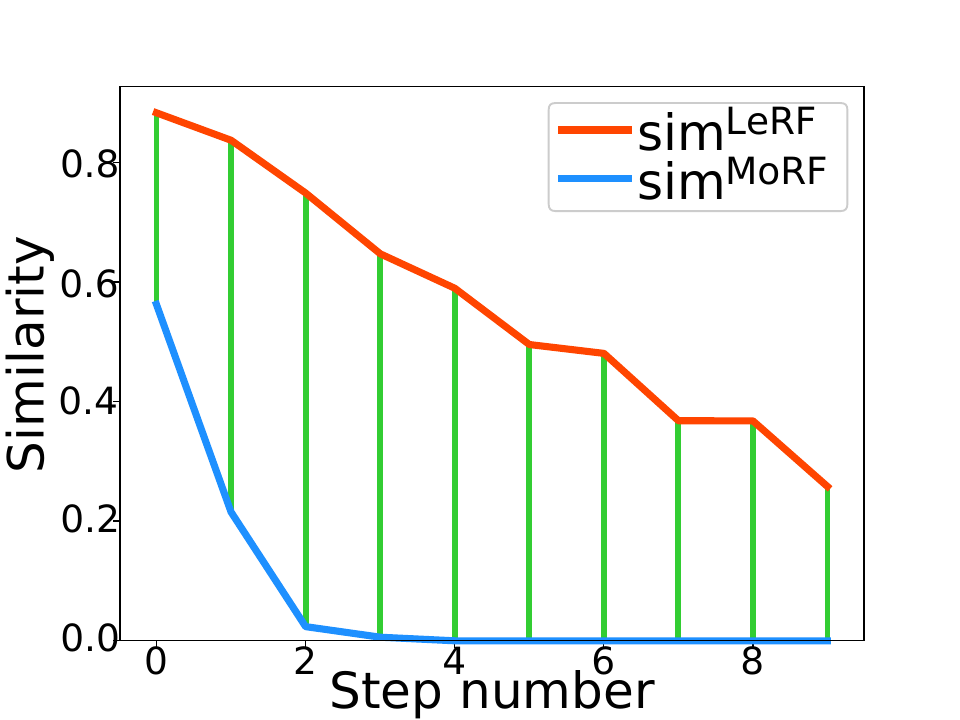}
         \Description{A graph with two line plots representing the attribution map similarity MoRF and LeRF change curves. Both curves are dropping down and to the right as more input neurons are masked. However, the MoRF curve drops significantly faster than the LeRF curve.}
         \caption{}
         \label{fig:morf_lerf_rel}
     \end{subfigure}
     \begin{subfigure}[b]{0.3\columnwidth}
         \centering
         \includegraphics[width=\textwidth]{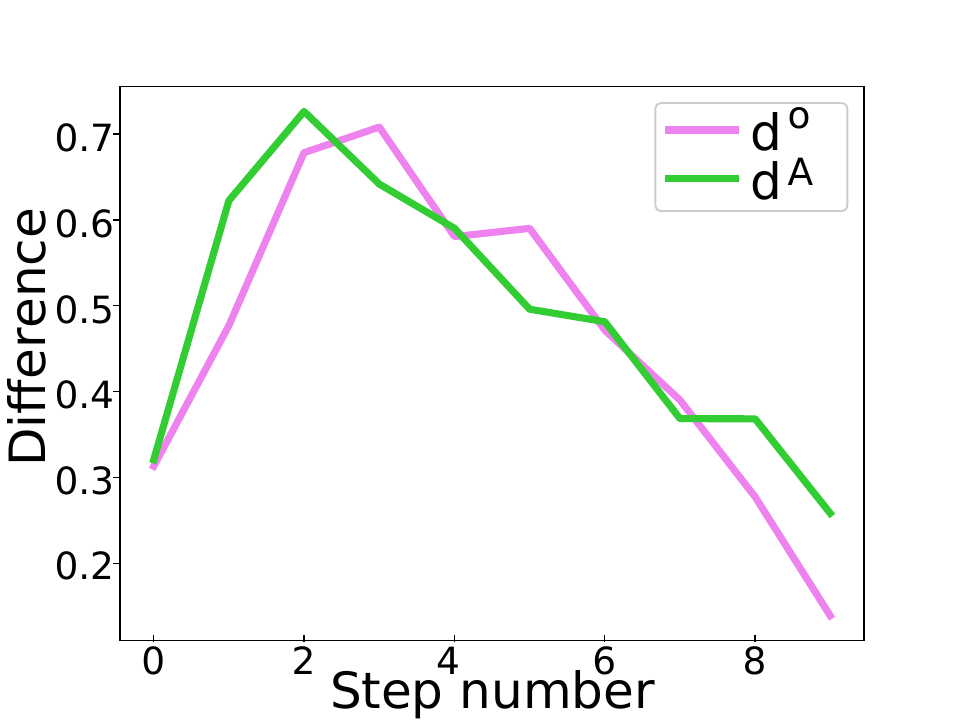}
         \Description{A graph with two line plots representing the per-step difference in MoRF and LeRF curves for the output and attribution map similarity as the input is being masked. Curves have a bell-like shape and are almost overlapping completely.}
         \caption{}
         \label{fig:diff_out_rel}
     \end{subfigure}

     \caption{Change curves for an example from ImageNet - VGG - absLRP: \textbf{(a)} MoRF and LeRF change curves for model output \textbf{(b)} MoRF and LeRF change curves for attribution maps \textbf{(c)} Difference curves for model output and attribution maps. Smaller difference in these curves signals a more locally robust method.}
        \label{fig:diff_curves}
\end{figure}

This produces a score between -1 and 1, indicating the quality of the local robustness of the attribution method. An example of the aforementioned curves obtained by our method on a random example from ImageNet can be seen in Figure \ref{fig:diff_curves}.

To evaluate the faithfulness aspect of the local consistency score, we make use of the recorded per-step maps of absolute products of inputs and gradients. Intuitively, when masking the least impactful features of the image, the highest values in these maps remain stable and centered around the classified object throughout the step-wise masking process. Conversely, when masking the most impactful features, these high impact values move across the image as they are progressively masked, resulting in the model receiving less and less evidence for the target class.

\begin{figure}
     \centering
      \begin{subfigure}[b]{0.18\columnwidth}
         \centering
         \includegraphics[width=\textwidth]{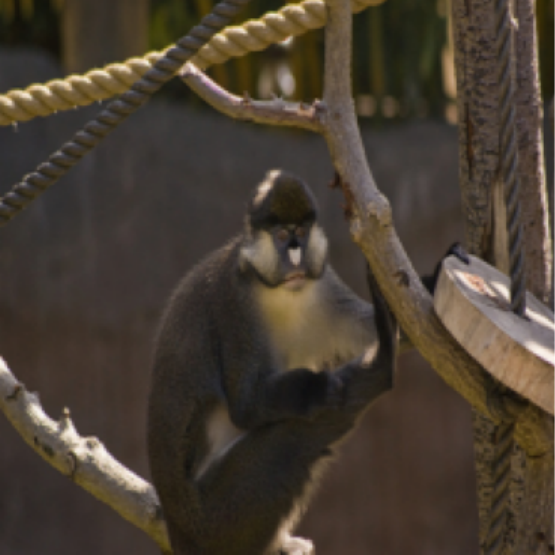}
        \Description{A photograph depicting a monkey standing on a branch.}
         \caption{}
         \label{fig:impact_original}
     \end{subfigure}
     \begin{subfigure}[b]{0.18\columnwidth}
         \centering
         \includegraphics[width=\textwidth]{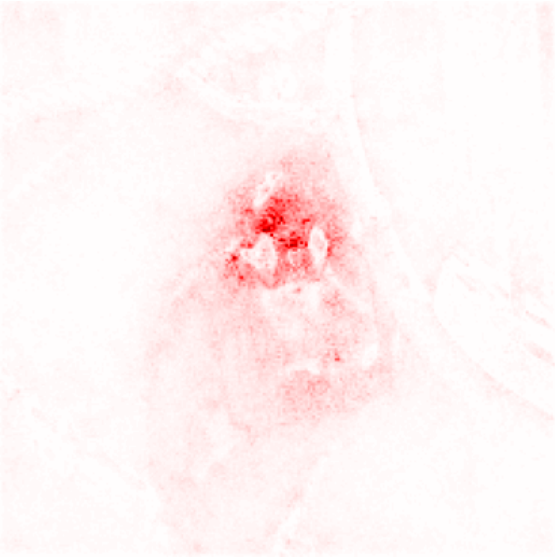}
         \Description{A one-channel image with its most intense pixels localized in the position of the monkey from the first figure.}
         \caption{}
         \label{fig:impact_back}
     \end{subfigure}
     \begin{subfigure}[b]{0.18\columnwidth}
         \centering
         \includegraphics[width=\textwidth]{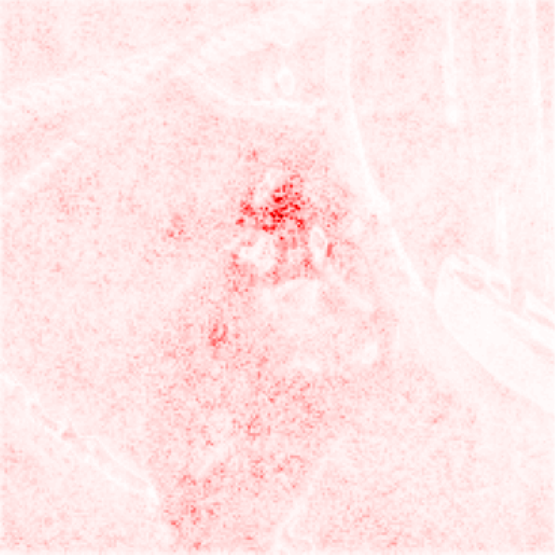}
         \Description{A one-channel image with a mostly uniform intensity of pixels throughout the image.}
         \caption{}
         \label{fig:impact_fore}
     \end{subfigure}
     \begin{subfigure}[b]{0.18\columnwidth}
         \centering
         \includegraphics[width=\textwidth]{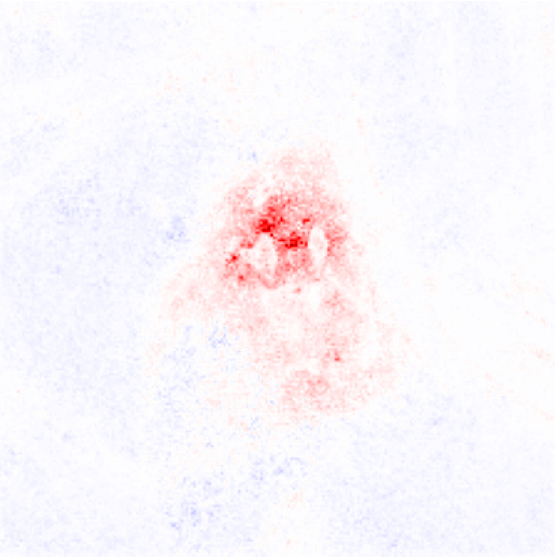}
         \Description{A one-channel image where the positive values are depicted in red and negative in blue. Red pixels are localized in the position of the monkey from the first figure. Blue pixels are localized in the background areas of the first figure. The most intense red pixels are centered around the monkey's head.}
         \caption{}
         \label{fig:impact_combined}
     \end{subfigure}
     \begin{subfigure}[b]{0.18 \columnwidth}
         \centering
         \includegraphics[width=\textwidth]{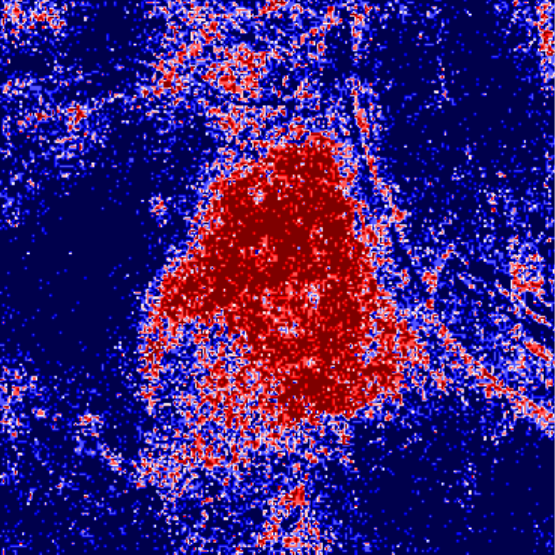}
         \Description{A one-channel image where the positive values are depicted in red and negative in blue. Red pixels are localized in the position of the monkey from the first figure. Blue pixels are localized in the background areas of the first figure. Because this figure is a sign of the previous figure, the positive areas are uniformly aligned with the monkey, while the negative areas uniformly cover the background.}
         \caption{}
         \label{fig:impact_sign}
     \end{subfigure}

     \caption{Impact maps on an example from ImageNet - VGG network: \textbf{(a)} Original image \textbf{(b)} Sum of per step absolute products of input and gradient when masking low impact features first \textbf{(c)} Masking high impact features first \textbf{(d)} Difference between (a) and (b) \textbf{(e)} Sign of (d)}
        \label{fig:impact_maps}
\end{figure}

As a result, the cumulative sum of these step-wise maps, obtained by masking the least impactful areas first, tends to be concentrated on the target object (Figure \ref{fig:impact_back}). Conversely, cumulative sums of maps obtained from masking the most impactful areas exhibit less focus on the target area and a more uniform distribution across the image background (Figure \ref{fig:impact_fore}). The disparity between these two maps yields the final map, referred to as the "combined impact map" ($I_c$) (Figure \ref{fig:impact_combined}), where positive values indicate high impact areas and negative values indicate low impact areas. This map is employed to evaluate the faithfulness of the attribution methods under assessment. By multiplying the sign of the combined impact map with the attribution, we differentiate correctly assessed high impact areas (positive attributing score and positive impact sign) from incorrectly assessed areas, where the attribution gave a high relevance score, but the impact sign was negative.

The faithfulness component of the local consistency score is defined as follows:

\begin{equation}
    \textbf{LC}_F(m_A) = \text{sum}\{\frac{A_{init} \cdot \text{sign}(I_c)}{|| A_{init} ||_1}\}
\end{equation}

where $\text{sum}$ denotes the sum over all dimensions of the tensor.

The sign of the impact map (Figure \ref{fig:impact_sign}) is multiplied by the attribution map and simply summed up and normalized by the initial sum of the attribution map. This results in a score that is between -1 and 1 depending on the local faithfulness of the attribution map.

Finally, the \textbf{Local consistency score} is defined as the positively clamped mean of the robustness and faithfulness parts of the score:

\begin{equation}
    \textbf{LC}(m_A) = (\frac{\text{LC}_R(m_A) + \text{LC}_F(m_A)}{2})^+
\end{equation}

The Local consistency factor in our metric effectively evaluates an attribution method's capability to identify highly influential regions within an image. However, it's important to note that these influential areas are not necessarily specific to a particular class, as multiple target classes can share the same influential regions. To address this limitation and provide a measure of the quality of class-specific influential areas, we introduce an additional factor in our metric called Contrastiveness.

\textbf{Contrastiveness} - inspired by the Focus metric \cite{arias2022focus}, our approach diverges by using four images—one positive and three negatives, alongside their predicted respective classes ($c_p$ for the positive and $c_{n_i}$ for the negatives). As there is only one positive and three negative quadrants, the labels need not be known in advance in order to sample two positives, as was the case in Focus. After obtaining softmax prediction scores $s^{p}$ from the model for the positive image, all four images are combined into a mosaic. The locations of the positive and negative images in the mosaic are randomly chosen. Next we define a \textit{scoring map} $S_{mosaic}$ with the same shape as the input mosaic, with the values inside the positive quadrant set to $1$, and inside other quadrants, indexed by $i$ as $\frac{2 s^{p}[c_{n_i}]}{s^{p}[c_p]} - 1$. This sets the values in negative quadrants as -1 if its respective negative class is not found in the positive image and conversely, for closer matches or identical classes, values approach 1. An extreme case of this is sampling multiple images with the same predicted class. In this case, all quadrants containing that class would be set to 1 if their class was chosen as the target. This way, we do not penalize the attribution method if a portion of its attribution falls into the quadrant containing similar or matching classes. Finally, we obtain the mosaic attribution map $A_{mosaic}$ for the positive target class and calculate the \textbf{Contrastiveness score}:

\begin{equation}
    \textbf{C}(m_A) = \text{sum}\{\frac{A_{mosaic} \cdot S_m}{|| A_{mosaic} ||_1}\}^+
\end{equation}

\begin{figure}
    \centering
     \begin{subfigure}[b]{0.25\columnwidth}
         \centering
         \includegraphics[width=\textwidth]{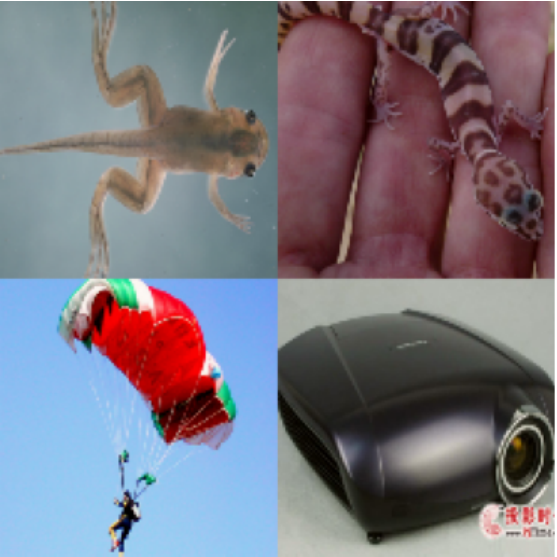}
         \Description{A mosaic made up of four images, one of them is a parachute that is in the bottom-left corner of the mosaic.}
         \caption{}
         \label{fig:mosaic_image}
     \end{subfigure}
     \begin{subfigure}[b]{0.25\columnwidth}
         \centering
         \includegraphics[width=\textwidth]{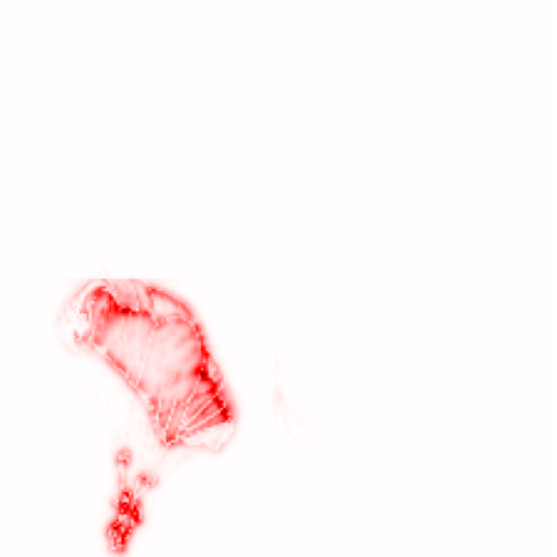}
         \Description{Attribution map of the previous image. The attribution is localized only on the parachute.}
         \caption{}
         \label{fig:mosaic_relevance}
     \end{subfigure}
     \begin{subfigure}[b]{0.25\columnwidth}
         \centering
         \includegraphics[width=\textwidth]{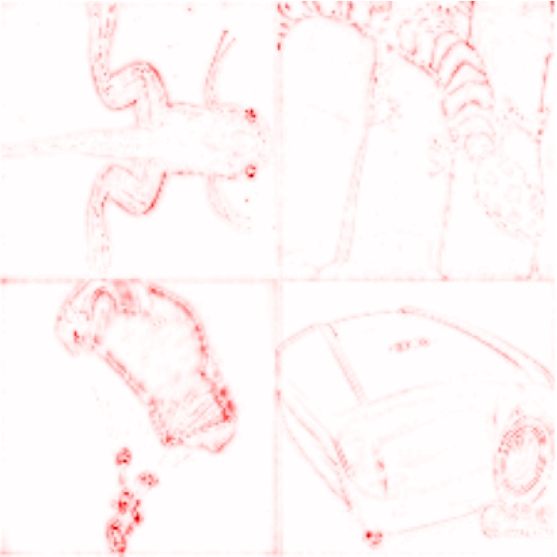}
         \Description{Attribution map of the first image. The attribution is uniformly spread among all four sub-images.}
         \caption{}
         \label{fig:mosaic_relevance_a2b1}
     \end{subfigure}
     \caption{\textbf{(a)} Example mosaic image - ImageNet \textbf{(b)} Our method's attribution map (positive only) - selected bottom left image as target class \textbf{(c)} LRP-$\alpha_2\beta_1$ attribution map (positive only), example of an attribution method with poor contrastive properties}
    \label{fig:mosaic}
\end{figure}

Contrastiveness scores close to 1 indicate a better ability to produce contrastive attribution maps. Examples of good and poor contrastiveness scores for the mosaic image in Figure \ref{fig:mosaic_image} are presented in Figures \ref{fig:mosaic_relevance} - our method and \ref{fig:mosaic_relevance_a2b1} - LRP-$\alpha_2\beta_1$ method. \\

Ultimately, in order to fulfill the requirements of Local consistency and Contrastiveness, a reliable attribution method should exhibit both factors. Hence, the final score of GAE for an attribution method $m_A$ is determined as the product of the two factors:

\begin{equation}
    \textbf{GAE}(m_A) = \text{LC}(m_A) \cdot \text{C}(m_A)
\end{equation}

In all the aforementioned calculations, we utilize the positive parts of the attribution maps, which are normalized so that the largest value corresponds to 1. By focusing on the positive parts and normalizing the maps, we ensure consistency and comparability in our calculations, allowing us to accurately evaluate the relevance and impact of different features in the attribution maps.

\section{Experiments} \label{sec:experiments}
In the Experiments section, we employ our novel metric to comprehensively evaluate the performance of our attribution method, alongside state-of-the-art and commonly used methods, across various models and datasets. To provide a comprehensive analysis, we include the results obtained from existing evaluation metrics in the domains of faithfulness, robustness, and localization. Additionally, we conduct an ablation study specifically for Vision Transformers, aiming to elucidate the key advantages of our attribution method compared to existing approaches. Furthermore, we conduct qualitative experiments to visually demonstrate the benefits of our attribution method in diverse scenarios. Finally, we showcase that our evaluation metric effectively distinguishes between high-quality explanations and low-quality explanations.
\subsection{Quantitative experiments}
We evaluate our proposed LRP method on two publicly available image classification datasets - ImageNet (Deng et al. \cite{deng2009imagenet}) and PascalVOC2012 (Everingham et al. \cite{Everingham10}), using the GAE evaluation metric. Step number for the local consistency masking procedure is set to $T=10$ in all experiments. For both datasets, we assess various attribution methods across three pre-trained image classification models: VGG16 \cite{NIPS2012_c399862d}, ResNet50 \cite{resnet} and ViT-Base \cite{vision_transformer}. The significance of the difference in the mean values of the total scores was verified using the Wilcoxon signed-rank test with $p=0.05$.

In the first quantitative experiment, the VGG network, we compare our approach to commonly used, and state-of-the-art explanation approaches for pure convolutional networks: Saliency \cite{https://doi.org/10.48550/arxiv.1312.6034}, Input*Gradient \cite{gradxinput}, Deconvolution \cite{https://doi.org/10.48550/arxiv.1311.2901}, LRP-$\epsilon$, DeepLIFT \cite{gradxinput}, LRP-$\alpha_1 \beta_0$, LRP-$\alpha_2 \beta_1$, Integrated Gradients \cite{https://doi.org/10.48550/arxiv.1703.01365}, SmoothGrad \cite{smilkov2017smoothgrad}, GradCAM \cite{Selvaraju_2019}, HiResCAM \cite{draelos2021use}, LayerCAM \cite{jiang2021layercam}, GradCAM++ \cite{chattopadhay2018grad}, contrastive-LRP (cLRP) \cite{gu2019understanding}, GuidedGradCAM \cite{Selvaraju_2019}, Relative Attributing Propagation (RAP) \cite{nam2019relative} and contrastive-RAP (cRAP) (an augmented version of RAP using the same method of obtaining contrastive maps described in contrastive-LRP paper). We also include two baseline attribution methods: a constant attribution that always produces ones as the attribution map and a random attribution sampled from a normal distribution.

The results of the first experiment can be found in Tables \ref{tab:vgg_gae_imagenet} and \ref{tab:vgg_gae_pascalvoc}. Our method outperforms all the aforementioned methods by a significant margin on both datasets, obtaining a total score of 0.272 on ImageNet and 0.238 on PascalVOC. It is followed by GuidedGradCAM with scores of 0.207 and 0.167 on the respective datasets, and cLRP with scores 0.12 and 0.112. The majority of the other methods score significantly lower than these methods, primarily because of their lack of contrastiveness, as their attribution maps do not vary depending on the chosen target class. This observation was also noted by Gu et al. \cite{gu2019understanding} and Chefer et al. \cite{beyond_transformer}. For completeness, we include the average values of Local consistency and Contrastiveness scores. While observing these scores separately may offer some insight into the problems or advantages of an attribution method, relying on only one factor will lead to similar pitfalls as previous evaluation methods.


\begin{table}
    \begin{minipage}{.49\textwidth}
    \captionof{table}{GAE scores obtained by various attribution \\
    methods on ImageNet using VGG architecture}
      \centering
      \begin{tabular}{cccc}
        \toprule
        Method name & $\overline{LC}$ & $\overline{C}$ & $\overline{Total}$ \\ 
         \midrule
        Constant & 0.0 & 0.0 & 0.0 \\
        Random & 0.0 & 0.0 & 0.0 \\
        Saliency & 0.0 & 0.011 & 0.0 \\
        Input*Gradient & 0.076 & 0.016 & 0.0 \\
        Deconvolution & 0.0 & 0.0 & 0.0 \\
        LRP-$\epsilon$ & 0.117 & 0.022 & 0.001 \\
        DeepLIFT & 0.159 & 0.08 & 0.011 \\
        LRP-$\alpha_1\beta_0$ & 0.318 & 0.003 & 0.001 \\
        LRP-$\alpha_2\beta_1$ & 0.307 & 0.016 & 0.006 \\
        Integrated Gradients & 0.139 & 0.022 & 0.002 \\
        SmoothGrad & 0.004 & 0.016 & 0.0 \\
        GradCAM & 0.132 & 0.424 & 0.06 \\
        HiResCAM & 0.107 & 0.342 & 0.038 \\
        LayerCAM & 0.079 & 0.089 & 0.008 \\
        GradCAM++ & 0.093 & 0.056 & 0.006 \\
        GuidedGradCAM & 0.281 & 0.665 & 0.207 \\
        cLRP & 0.3 & 0.407 & 0.12 \\
        RAP & 0.383 & 0.202 & 0.097 \\
        cRAP & 0.106 & 0.542 & 0.053 \\
        \textbf{Ours} & 0.396 & 0.676 & \textbf{0.272} \\
         \bottomrule 
      \end{tabular}
      \label{tab:vgg_gae_imagenet}
    \end{minipage}
    \begin{minipage}{.49\textwidth}
    \captionof{table}{GAE scores obtained by various attribution \\ methods on PascalVOC using VGG architecture}
      \centering
      \begin{tabular}{cccc}
        \toprule
        Method name & $\overline{LC}$ & $\overline{C}$ & $\overline{Total}$ \\ 
         \midrule
        Constant & 0.0 & 0.001 & 0.0 \\
        Random & 0.0 & 0.001 & 0.0 \\
        Saliency & 0.0 & 0.013 & 0.0 \\
        Input*Gradient & 0.085 & 0.023 & 0.002 \\
        Deconvolution & 0.0 & 0.002 & 0.0 \\
        LRP-$\epsilon$ & 0.134 & 0.027 & 0.003 \\
        DeepLIFT & 0.161 & 0.071 & 0.01 \\
        LRP-$\alpha_1\beta_0$ & 0.312 & 0.009 & 0.002 \\
        LRP-$\alpha_2\beta_1$ & 0.3 & 0.022 & 0.006 \\
        Integrated Gradients & 0.157 & 0.029 & 0.004 \\
        SmoothGrad & 0.002 & 0.018 & 0.0 \\
        GradCAM & 0.112 & 0.372 & 0.042 \\
        HiResCAM & 0.096 & 0.287 & 0.025 \\
        LayerCAM & 0.069 & 0.089 & 0.007 \\
        GradCAM++ & 0.093 & 0.063 & 0.006 \\
        GuidedGradCAM & 0.254 & 0.606 & 0.167 \\
        cLRP & 0.267 & 0.398 & 0.112 \\
        RAP & 0.36 & 0.242 & 0.094 \\
        cRAP & 0.11 & 0.497 & 0.054 \\
        \textbf{Ours} & 0.399 & 0.594 & \textbf{0.238} \\
         \bottomrule 
      \end{tabular}
      \label{tab:vgg_gae_pascalvoc}
    \end{minipage}
  \end{table}

In the subsequent experiment, we assess the performance of several attribution methods, including Integrated Gradients, SmoothGrad, GradCAM, HiResCAM, LayerCAM, GradCAM++, cLRP, GuidedGradCAM, RAP, cRAP, and our newly proposed method. These methods are evaluated using the ResNet50 model on both the ImageNet and PascalVOC datasets.

As can be seen from Tables \ref{tab:resnet50_gae_imagenet} and \ref{tab:resnet50_gae_pascalvoc}, our proposed method consistently demonstrates superior performance compared to the majority of other evaluated methods. However, it is worth noting that the cLRP method exhibits the same level of performance as our method for this particular model. We hypothesize that this could be attributed to the model's reduced susceptibility to the issue of relative magnitude attribution, as depicted in Figure \ref{fig:intline_vs_alpha1beta0}. Nonetheless, the precise reason behind this susceptibility remains unknown, as it could potentially be influenced by various factors such as the residual architecture, batch normalization, or differences in training hyperparameters.

Remarkably, both our proposed method and the cLRP method outperform all other methods by a significant margin, with their respective total GAE scores being more than double that of the next-best attribution method, which in this case is GuidedGradCAM.

\begin{table}
    \begin{minipage}{.49\textwidth}
    \captionof{table}{GAE scores obtained by various attribution \\ methods on ImageNet using ResNet50 architecture.}
      \centering
      \begin{tabular}{cccc}
        \toprule
        Method name & $\overline{LC}$ & $\overline{C}$ & $\overline{Total}$ \\ 
         \midrule
        Integrated Gradients & 0.127 & 0.119 & 0.015 \\
        SmoothGrad & 0.003 & 0.141 & 0.001 \\
        GradCAM & 0.181 & 0.385 & 0.07 \\
        HiResCAM & 0.186 & 0.307 & 0.056 \\
        LayerCAM & 0.187 & 0.299 & 0.055 \\
        GradCAM++ & 0.24 & 0.391 & 0.093 \\
        GuidedGradCAM & 0.243 & 0.523 & 0.127 \\
        cLRP & 0.358 & 0.792 & \textbf{0.283}* \\
        RAP & 0.004 & 0.109 & 0.001 \\
        cRAP & 0.073 & 0.599 & 0.042 \\
        \textbf{Ours} & 0.361 & 0.75 & \textbf{0.272}* \\ 
         \bottomrule 
      \end{tabular}
      \footnotesize *Difference in results could not be confirmed \\
      by Wilcoxon signed-rank test at p=0.05
    \label{tab:resnet50_gae_imagenet}
    \end{minipage}
    \begin{minipage}{.49\textwidth}
    \captionof{table}{GAE scores obtained by various attribution \\ methods on PascalVOC using ResNet50 architecture.}
      \centering
      \begin{tabular}{cccc}
        \toprule
        Method name & $\overline{LC}$ & $\overline{C}$ & $\overline{Total}$ \\ 
         \midrule
        Integrated Gradients & 0.088 & 0.101 & 0.009 \\
        SmoothGrad & 0.0 & 0.128 & 0.0 \\
        GradCAM & 0.165 & 0.391 & 0.07 \\
        HiResCAM & 0.165 & 0.299 & 0.054 \\
        LayerCAM & 0.157 & 0.261 & 0.047 \\
        GradCAM++ & 0.21 & 0.371 & 0.082 \\
        GuidedGradCAM & 0.189 & 0.576 & 0.11 \\
        cLRP & 0.312 & 0.863 & \textbf{0.266}* \\
        RAP & 0.005 & 0.099 & 0.002 \\
        cRAP & 0.058 & 0.597 & 0.04 \\
        \textbf{Ours} & 0.299 & 0.828 & \textbf{0.251}* \\ 
         \bottomrule 
      \end{tabular}
      \footnotesize *Difference in results could not be confirmed \\
      by Wilcoxon signed-rank test at p=0.05
      \label{tab:resnet50_gae_pascalvoc}
    \end{minipage}
  \end{table}

Lastly, we assess the performance of various attribution methods for Vision Transformers, including GradCAM, GradCAM++, HiResCAM, LayerCAM, Last Layer Attention, Rollout, Transformer Interpretability Beyond Attention Visualization (TIBAV), and our proposed method.

Consistent with previous experiments, our method outperforms all other attribution methods by a significant margin and achieves the highest score on the new evaluation method. The results are presented in Tables \ref{tab:vit_gae_imagenet} and \ref{tab:vit_gae_pascalvoc}. When examining the mean Local consistency scores, our method demonstrates a score that is more than double that of the subsequent method, HiResCAM. However, gradient-based methods exhibit an advantage in the Contrastiveness score, consistently outperforming our method in this aspect. It is noteworthy that Rollout and Last Layer Attention, as noted by Chefer et al. \cite{beyond_transformer}, do not generate contrastive maps, resulting in consistently lower scores on the Contrastiveness metric. TIBAV, being a combination of gradient-based and Rollout methods, falls in-between their respective performances for each of the scores.

It is important to acknowledge the relatively lower score of our method in the Vision Transformer experiment compared to the two previous experiments. This may suggest ample room for improvement in our method by defining different rules for the Transformer-specific layers or defining different contrastive propagation rules.

\begin{table}
    \begin{minipage}{.49\textwidth}
    \captionof{table}{GAE scores obtained by various attribution \\ methods on ImageNet using ViT-Base patch \\ size 16 architecture.}
      \centering
      \begin{tabular}{cccc}
        \toprule
        Method name & $\overline{LC}$ & $\overline{C}$ & $\overline{Total}$ \\ 
         \midrule
        GradCAM & 0.042 & 0.651 & 0.024 \\
        GradCAM++ & 0.003 & 0.164 & 0.0 \\
        HiResCAM & 0.06 & 0.821 & 0.047 \\
        LayerCAM & 0.032 & 0.745 & 0.022 \\
        LLAttention & 0.002 & 0.0 & 0.0 \\
        Rollout & 0.005 & 0.022 & 0.0 \\
        TIBAV & 0.018 & 0.475 & 0.006 \\
        \textbf{Ours} & 0.162 & 0.564 & \textbf{0.075} \\ 
         \bottomrule 
      \end{tabular}
      \label{tab:vit_gae_imagenet}
    \end{minipage}
    \begin{minipage}{.49\textwidth}
    \captionof{table}{GAE scores obtained by various attribution \\ methods on PascalVOC using ViT-Base patch \\ size 16 architecture.}
      \centering
      \begin{tabular}{cccc}
        \toprule
        Method name & $\overline{LC}$ & $\overline{C}$ & $\overline{Total}$ \\ 
         \midrule
        GradCAM & 0.034 & 0.53 & 0.017 \\
        GradCAM++ & 0.002 & 0.169 & 0.0 \\
        HiResCAM & 0.072 & 0.679 & 0.045 \\
        LayerCAM & 0.042 & 0.557 & 0.02 \\
        LLAttention & 0.001 & 0.008 & 0.0 \\
        Rollout & 0.005 & 0.027 & 0.0 \\
        TIBAV & 0.02 & 0.394 & 0.008 \\
        \textbf{Ours} & 0.164 & 0.476 & \textbf{0.072} \\ 
         \bottomrule 
      \end{tabular}
      \label{tab:vit_gae_pascalvoc}
    \end{minipage}
  \end{table}

For completeness, we also evaluate all methods on both datasets using commonly used evaluation metrics for attribution-based methods described in Section \ref{rel_sec}: ROAD (Rong et al. \cite{2202.00449}), Local Lipschitz Estimate (Alvarez et al. \cite{alvarez2018robustness}) and the Focus metric (Arias-Duart et al. \cite{arias2022focus}).

\begin{table}
    \caption{Scores obtained by various attribution methods on other metrics on ImageNet dataset using VGG architecture.}
    \label{tab:vgg_results_other_imagenet}
    \begin{minipage}{\columnwidth}
    \begin{center}
    \begin{tabular}{c c c c c}
        \toprule
        Method name & $\overline{ROAD^{MoRF} (\downarrow)}$ & $\overline{ROAD^{LeRF} (\uparrow)}$ &$\overline{Lipschitz (\downarrow)}$ & $\overline{Focus (\uparrow)}$ \\ 
         \midrule
        Constant & 0.461 & 0.459 & \textbf{0.0} & 0.5 \\
        Random & 0.585 & 0.587 & 1.123 & 0.5 \\
        Saliency & 0.398 & 0.65 & 0.459 & 0.549 \\
        Input*Gradient & 0.352 & 0.619 & 0.371 & 0.547 \\
        Deconvolution & 0.483 & 0.639 & 0.445 & 0.501 \\
        LRP-$\epsilon$ & 0.3 & 0.678 & 0.341 & 0.566 \\
        DeepLIFT & 0.244 & 0.731 & 0.291 & 0.621 \\
        LRP-$\alpha_1\beta_0$ & \textbf{0.199}* & 0.762 & 0.126 & 0.506 \\
        LRP-$\alpha_2\beta_1$ & 0.226 & 0.776 & 0.087 & 0.535 \\
        Integrated Gradients & 0.302 & 0.669 & 0.331 & 0.559 \\
        SmoothGrad & 0.28 & 0.746 & 0.357 & 0.561 \\
        GradCAM & 0.207 & \textbf{0.798}* & 0.687 & 0.844 \\
        HiResCAM & 0.204 & 0.777 & 0.634 & 0.807 \\
        LayerCAM & \textbf{0.199}* & 0.779 & 0.499 & 0.642 \\
        GradCAM++ & 0.213 & 0.758 & 0.496 & 0.628 \\
        GuidedGradCAM & \textbf{0.177}* & \textbf{0.818}* & 0.13 & \textbf{0.886}* \\
        cLRP & 0.229 & 0.79 & 0.167 & 0.833 \\ 
        RAP & 0.216 & 0.77 & 0.176 & 0.482 \\ 
        cRAP & 0.302 & 0.702 & 0.401 & 0.833 \\ 
        \textbf{Ours} & \textbf{0.184}* & \textbf{0.792}* & 0.23 & \textbf{0.885}* \\ 
         \bottomrule
    \end{tabular}
\end{center}
\end{minipage}
\footnotesize $\downarrow$ - lower score is better; $\uparrow$ - higher score is better \\
\footnotesize *Difference in results could not be confirmed by Wilcoxon signed-rank test at p=0.05
\end{table}

\begin{table}
    \caption{Scores obtained by various attribution methods on other metrics on ImageNet dataset using ResNet50 architecture.}
    \label{tab:resnet50_results_other_imagenet}
    \begin{minipage}{\columnwidth}
    \begin{center}
    \begin{tabular}{c c c c c}
        \toprule
        Method name & $\overline{ROAD^{MoRF} (\downarrow)}$ & $\overline{ROAD^{LeRF} (\uparrow)}$ &$\overline{Lipschitz (\downarrow)}$ & $\overline{Focus (\uparrow)}$ \\ 
         \midrule
        Integrated Gradients & 0.452 & 0.773 & 0.338 & 0.65 \\
        SmoothGrad & 0.416 & \textbf{0.856}* & 0.336 & 0.665 \\
        GradCAM & 0.298 & 0.792 & 0.417 & 0.769 \\
        HiResCAM & 0.297 & 0.787 & 0.42 & 0.714 \\
        LayerCAM & 0.305 & 0.777 & 0.457 & 0.685 \\
        GradCAM++ & 0.311 & 0.777 & 0.479 & 747 \\
        GuidedGradCAM & 0.305 & \textbf{0.857}* & \textbf{0.126}* & 0.803 \\
        cLRP & \textbf{0.25}* & \textbf{0.866}* & \textbf{0.13}* & \textbf{0.96}* \\
        RAP & 0.42 & 0.689 & 0.814 & 0.544 \\
        cRAP & 0.413 & 0.759 & 0.744 & 0.878 \\
        \textbf{Ours} & \textbf{0.266}* & 0.836 & 0.243 & \textbf{0.966}* \\ 
         \bottomrule
    \end{tabular}
\end{center}
\end{minipage}
\footnotesize $\downarrow$ - lower score is better; $\uparrow$ - higher score is better \\
\footnotesize *Difference in results could not be confirmed by Wilcoxon signed-rank test at p=0.05
\end{table}

\begin{table}
    \caption{Scores obtained by various attribution methods on other metrics - ViT-Base patch size 16 - ImageNet dataset}
    \label{tab:vitbase_results_other_imagenet}
    \begin{minipage}{\columnwidth}
    \begin{center}
    \begin{tabular}{c c c c c}
        \toprule
        Method name & $\overline{ROAD^{MoRF} (\downarrow)}$ & $\overline{ROAD^{LeRF} (\uparrow)}$ &$\overline{Lipschitz (\downarrow)}$ & $\overline{Focus (\uparrow)}$ \\ 
         \midrule
        GradCAM & \textbf{0.317} & \textbf{0.82}* & 0.867 & \textbf{0.913}* \\
        GradCAM++ & 0.56 & 0.678 & 2.556 & 0.707 \\
        HiResCAM & 0.383 & \textbf{0.833}* & 0.61 & \textbf{0.913}* \\
        LayerCAM & 0.399 & \textbf{0.819}* & 0.665 & 0.883 \\
        LLAttention & 0.406 & 0.774 & 0.54 & 0.472 \\
        Rollout & 0.414 & 0.789 & 1.064 & 0.496 \\
        TIBAV & 0.384 & \textbf{0.813}* & 0.876 & 0.798 \\
        \textbf{Ours} & 0.457 & \textbf{0.803}* & \textbf{0.477} & 0.826 \\ 
         \bottomrule
    \end{tabular}
\end{center}
\end{minipage}
\footnotesize $\downarrow$ - lower score is better; $\uparrow$ - higher score is better \\
\footnotesize *Difference in results could not be confirmed by Wilcoxon signed-rank test at p=0.05
\end{table}

The results obtained on the ImageNet dataset are presented in Table \ref{tab:vgg_results_other_imagenet} for VGG, Table \ref{tab:resnet50_results_other_imagenet} for ResNet50; and finally Table \ref{tab:vitbase_results_other_imagenet} for ViT-Base. Results of the experiments conducted on the PascalVOC dataset can be found in Appendix \ref{apx:pascalvoc_apx}. These tables demonstrate that absLRP consistently achieves top or near-top performance across various model-dataset settings, as reflected by most metrics. However, it is worth noting that in the ViT-Base experiments, we observed a lower Focus score, which aligns with the findings of the Contrastiveness score of GAE, suggesting the potential for exploring alternative methods of contrastive map calculation in this particular scenario.

Furthermore, it is crucial to highlight that while absLRP consistently excels on conventional metrics, the substantial performance gaps observed on the GAE metric are not as prominently reflected in these standardized measures. This discrepancy arises from inherent limitations in standardized metrics, as discussed in Section \ref{evaluation_metrics}. Additionally, Additionally, evaluating each property (faithfulness, robustness, and localization) in isolation overlooks the synergistic value derived from their combined assessment. An attribution method may score high in faithfulness and robustness for a specific image but exhibit poor localization. Conversely, a different image may skew the values towards localization while lacking in faithfulness or robustness. True excellence in an attribution method is realized only when all three properties are concurrently satisfied for each image.

\subsection{Ablation study}
We examine three ablated variants of our method specifically designed for Vision Transformers:

1. Patch-level propagation: In this variant, we propagate the attributions solely to the patch-level, disregarding the relevance to individual input pixels.

2. Values relevance: This variant involves propagating only the values' relevances through the self-attention block, while excluding the relevances of queries and keys. This approach aligns with the implementation of self-attention relevance propagation by Ali et al. \cite{xai_bert}.

3. Queries and keys relevance: In this variant, we focus on propagating only the relevance of queries and keys, neglecting the relevance of values. This mirrors the methodology of TIBAV, utilizing only the self-attention scores derived from the multiplication of queries and keys.

By exploring these ablated variants, we aim to investigate the impact of different levels of attribution propagation on the performance of our method for Vision Transformers.

\begin{table}
    \caption{Ablation study of our method - ViT-Base patch size 16}
    \label{tab:ablation_study}
    \begin{minipage}{\columnwidth}
    \begin{center}
    \begin{tabular}{c c c c c}
        \toprule
        & ImageNet & PascalVOC\\ 
         \midrule
        \textbf{Ours} & \textbf{0.075} & \textbf{0.072} \\ 
         Patch relevance & 0.041 (-45.3\%) & 0.035 (-51.4\%) \\
        Value relevance & 0.064 (-14.7\%) & 0.047 (-34.7\%)  \\
        Query and key relevance & 0.037 (-50.7\%) & 0.04 (-44.4\%) \\
         \bottomrule
    \end{tabular}
\end{center}
\end{minipage}
\end{table}

Table \ref{tab:ablation_study} displays the results of our ablation study on both datasets. Halting attribution propagation to input pixels significantly reduces our method's score. Notably, a substantial portion of the score is lost when only propagating query and key attributions through the network, which partly elucidates the relatively lower performance of TIBAV, which incorporates both principles. Exclusive propagation of value relevances through the self-attention block leads to a moderate score reduction. This ablation study strongly indicates that propagating relevance through all the operations within the Vision Transformer is crucial for obtaining high-quality attributions.

\subsection{Qualitative experiments}

Lastly, in this section, we present the qualitative results of our methods in various settings. The presented results are representative of the general performance of our methods.

\begin{figure}
  \centering
  \includegraphics[width=0.6\textwidth]{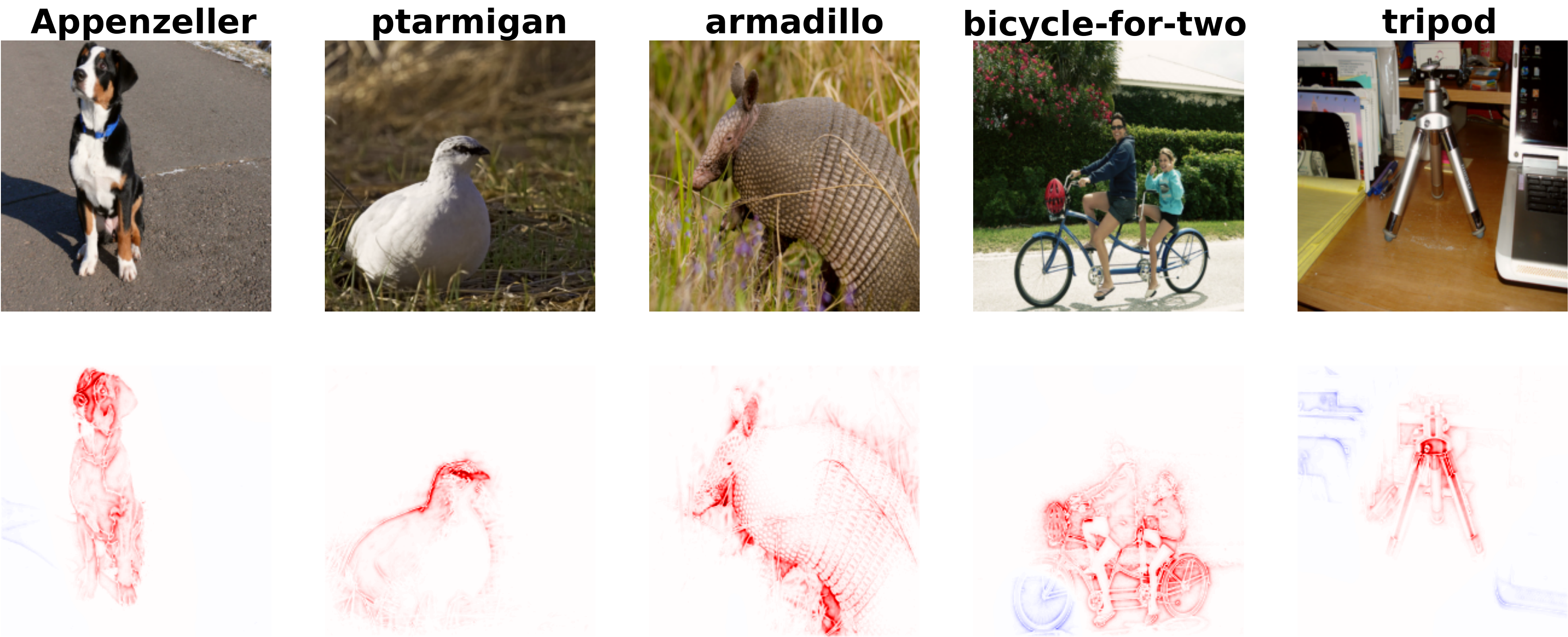}
  \Description{Two rows of five images. The first one depicts examples of different classes from the ImageNet dataset. The second row depicts attribution maps for the aforementioned images. Attributions are localized on the target object.}
  \captionof{figure}{absLRP attribution maps for VGG - ImageNet}
  \label{fig:vgg_simple_visualization}
\end{figure}

\begin{figure}
  \centering
  \includegraphics[width=0.6\textwidth]{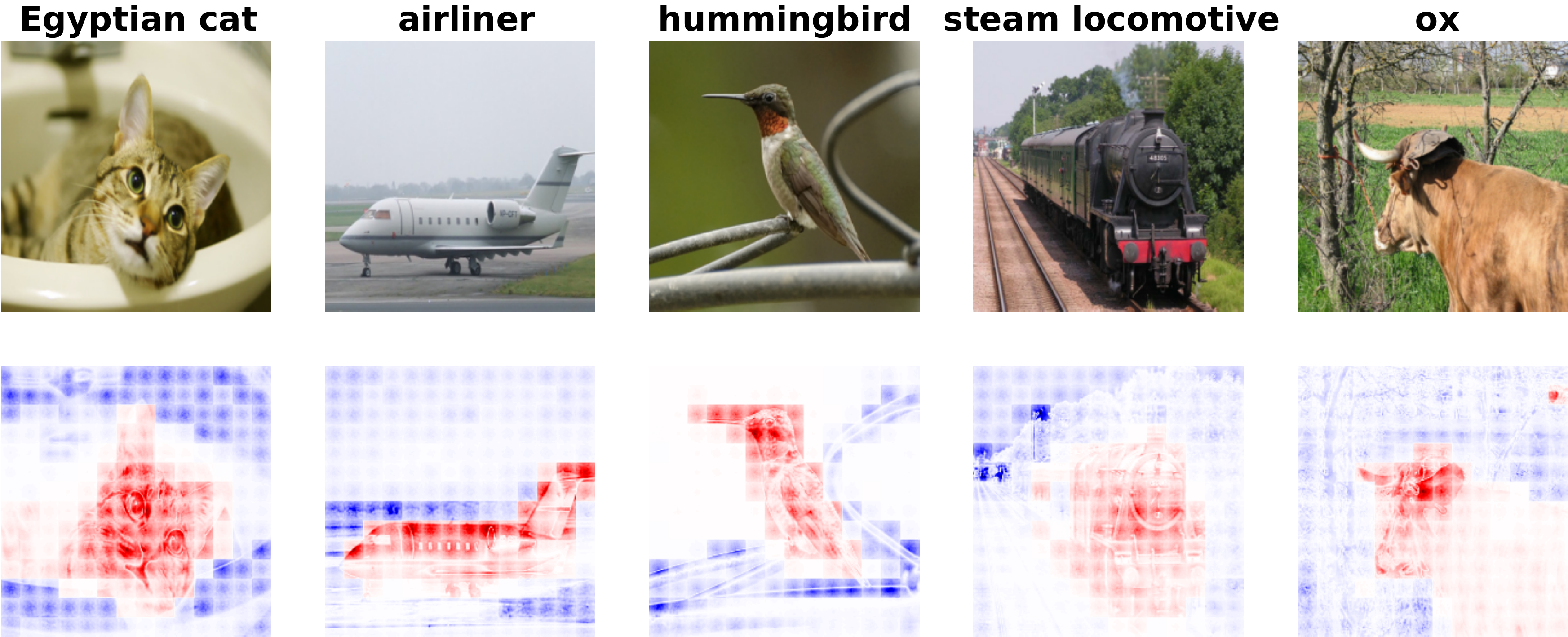}
  \Description{Two rows of five images. The first one depicts examples of different classes from the PascalVOC dataset. The second row depicts attribution maps for the aforementioned images. Attributions are localized on the target object.}
  \captionof{figure}{absLRP attribution maps for ViT-Base - PascalVOC}
  \label{fig:vit_simple_visualization}
\end{figure}

\begin{figure}
  \centering
  \includegraphics[height=5.5cm]{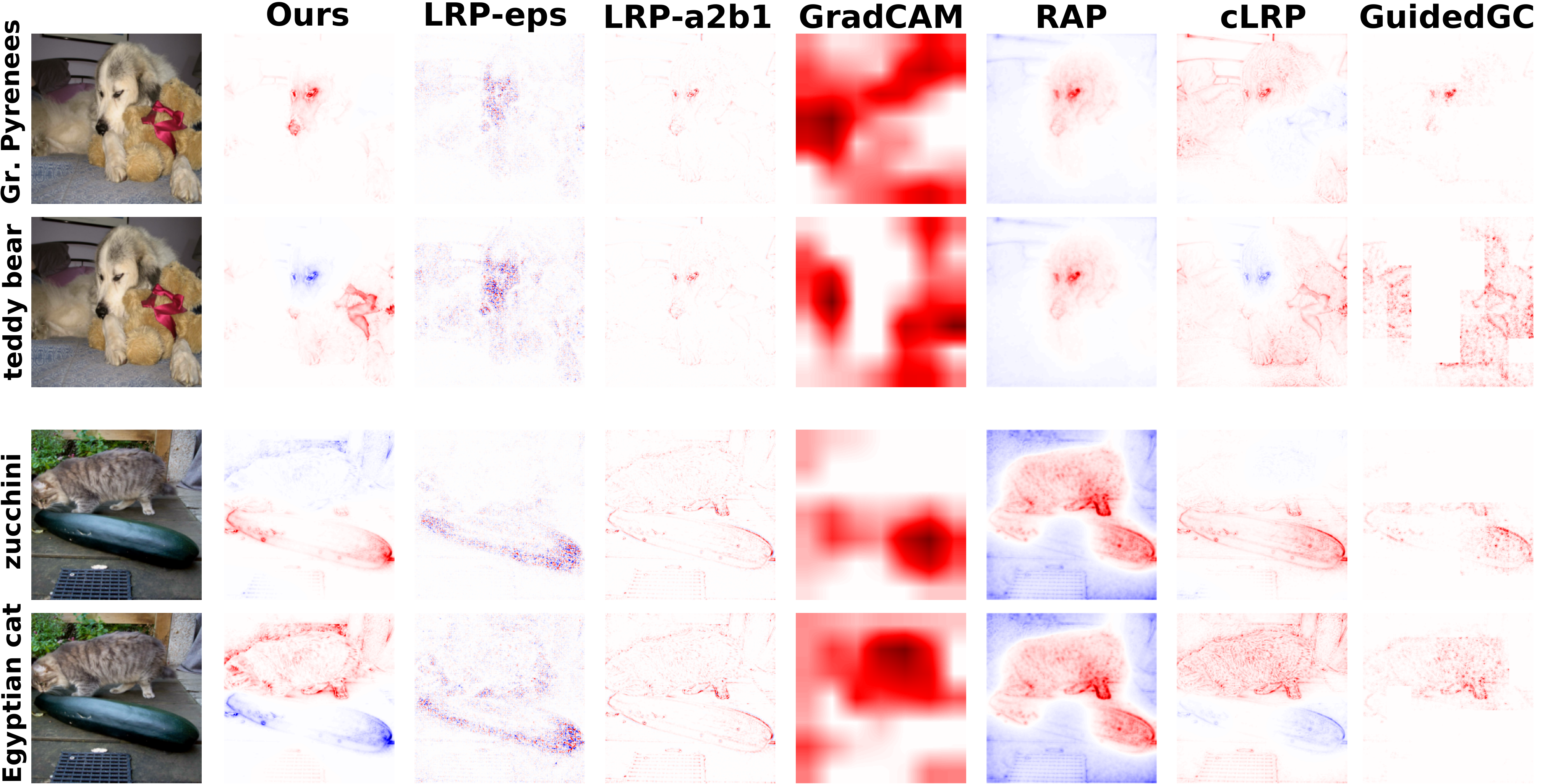}
  \Description{A grid of images and their respective attribution maps. Every pair of rows starts with the same image, but different target classes. In each row, the attributions of absLRP, LRP-alpha2beta1, LRP-epsilon, GradCAM, cRAP, cLRP and GuidedGradCAM are visualized.}
  \captionof{figure}{Class-specific visualizations for multiple attribution methods - ImageNet VGG}
  \label{fig:vgg_contrastive}
\end{figure}

\begin{figure}
  \centering
  \includegraphics[height=5.5cm]{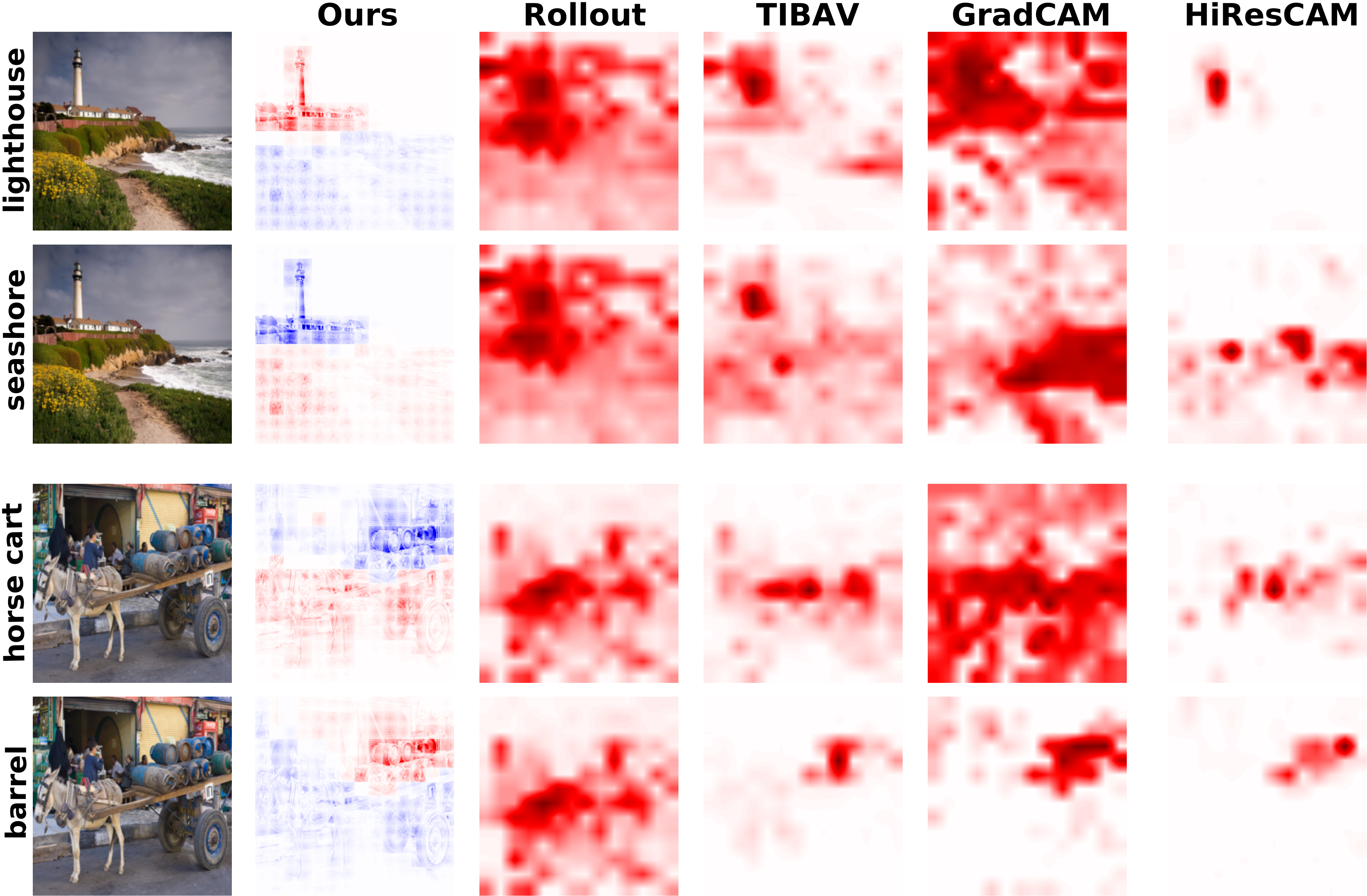}
  \Description{A grid of images and their respective attribution maps. Every pair of rows starts with the same image, but different target classes. In each row, the attributions of absLRP, Rollout, TIBAV, GradCAM and HiResCAM are visualized.}
  \captionof{figure}{Class-specific visualizations for multiple attribution methods - ImageNet ViT-Base}
  \label{fig:vit_contrastive}
\end{figure}

In the first qualitative experiment, we simply visualize the attribution maps of several randomly sampled images from the ImageNet and PascalVOC datasets for each of the evaluated models. The results for VGG and ViT-Base can be seen in Figures \ref{fig:vgg_simple_visualization} and \ref{fig:vit_simple_visualization} respectively. Alongside that, we present the attribution maps of several images containing multiple classes to show the contrastiveness property of our method, compared to several other methods from our quantitative tests. The results of this can be seen in Figure \ref{fig:vgg_contrastive} for VGG and \ref{fig:vit_contrastive} for ViT-Base on ImageNet dataset. Additional qualitative experiments for the VGG and ViT-Base models can be found in the Appendix \ref{apx:vgg_vis} and \ref{apx:vit_vis}. Qualitative experiments performed on the ResNet50 architecture can be found in the Appendix \ref{apx:resnet_vis}. 

\begin{figure}
\captionsetup[subfigure]{labelformat=empty}
    \centering
     \begin{subfigure}[b]{\columnwidth}
         \centering
         \includegraphics[height=0.3cm]{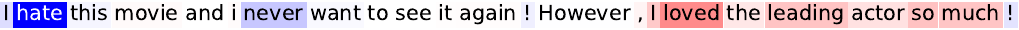}
         \Description{AbsLRP attribution map of a sentiment classification model. The movie review is: "I hate this movie and i never want to see it again! However, I loved the leading actor so much!" and the absLRP method highlights "I hate" in blue, and "I loved the leading actor so much" in red.}
     \end{subfigure}
     \begin{subfigure}[b]{\columnwidth}
         \centering
         \includegraphics[height=0.3cm]{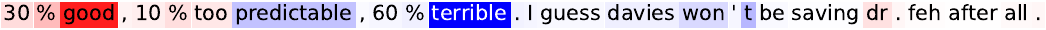}
         \Description{AbsLRP attribution map of a sentiment classification model. The movie review is: "30\% good, 10\% too predictable, 60\% terrible. I guess davies won't be saving dr. feh after all." and the absLRP method highlights "predictable" and "terrible" in blue; and "30\% good" in red.}
     \end{subfigure}
     \begin{subfigure}[b]{\columnwidth}
         \centering
         \includegraphics[height=0.3cm]{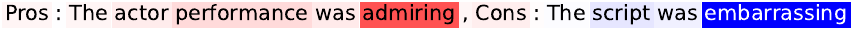}
         \Description{AbsLRP attribution map of a sentiment classification model. The movie review is: "Pros: The actor performance was admiring, Cons: The script was embarrassing" and the absLRP method highlights "script was embarrassing" in blue, and "performance was admiring" in red.}
     \end{subfigure}
     \caption{AbsLRP attribution maps for a BERT-based text sentiment classification model. Positive sentiment-contributing tokens are highlighted in red, and negative in blue.}
    \label{fig:distilbert_contrastive}
\end{figure}

For the ViT-Base attributions, our attributions keep the patchy structure caused by the model's architecture. However, it stands out as the only method that preserves discernible details within these patches, extending all the way down to the individual input pixels.

\begin{figure}
    \centering
    \includegraphics[width=0.6\textwidth]{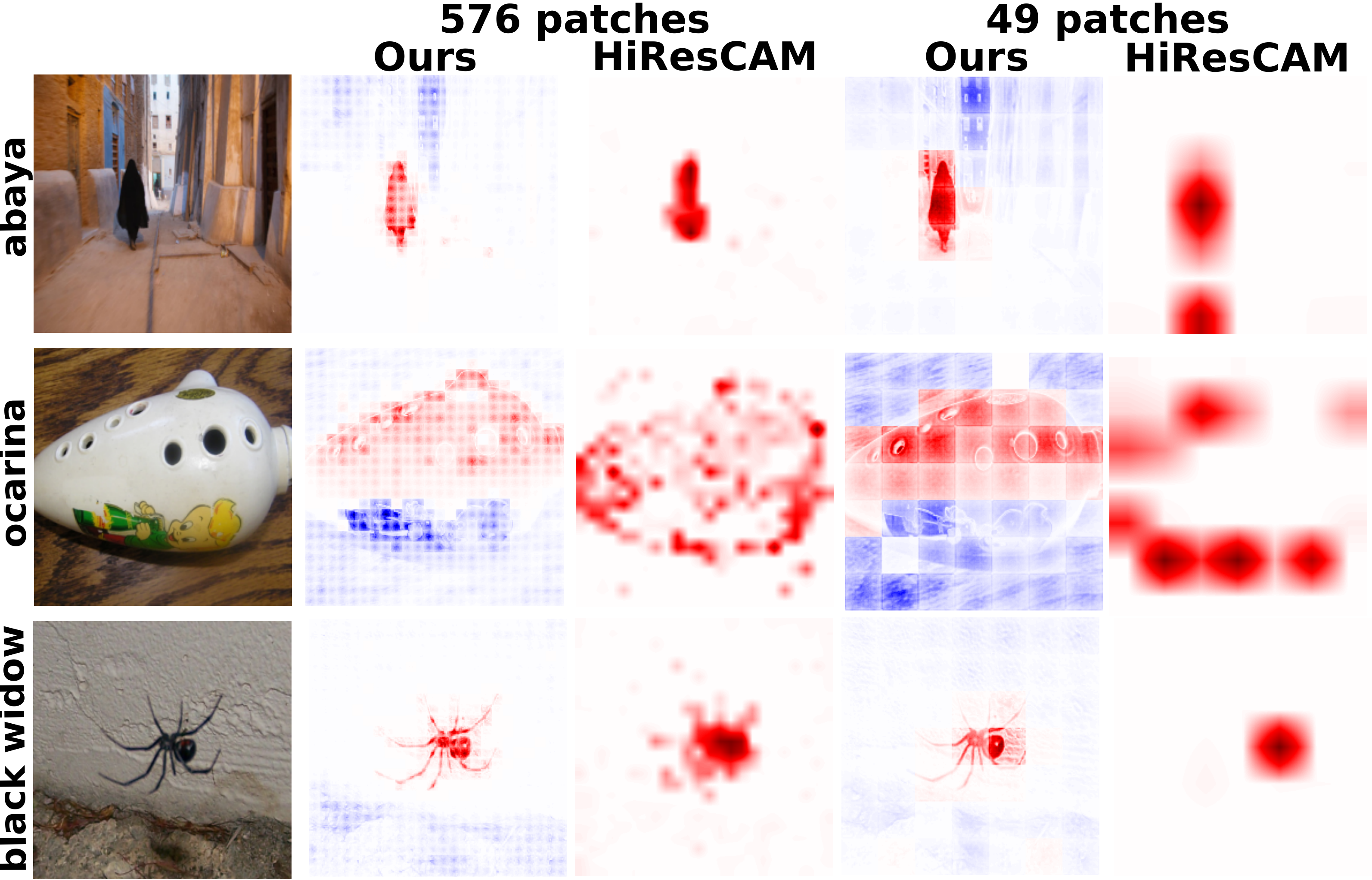}
    \Description{A three-by-five grid of images. In each row are firstly the original images from the ImageNet dataset. Depicted in the second and third column are the attribution maps of absLRP and HiResCAM methods for the model with the large number of patches. In the fourth and fifth column are the attribution maps for the model with the smaller number of patches. Level of fidelity of absLRP's attributions is in both cases pixel-level, while HiResCAM loses fidelity with smaller patch counts.}
    \caption{Attribution maps produced by absLRP and HiResCAM for ViT-Base patch size 16 image size 384 (576 patches) - first two columns; and ViT-Base patch size 32 image size 224 (49 patches) - last two columns. Our method provides pixel-level attributions even for models with a small patch count.}
    \label{fig:vit_base_patches}
\end{figure}

This brings us to the key advantage of our method compared to the other methods for Vision Transformer attribution - pixel level attribution. In Figure \ref{fig:vit_base_patches} we visualize the attribution maps for two different models, ViT-Base with patch sizes of 16x16 pixels, and input image size of 384x384 pixels, a total of 576 patches used; and ViT-Base with patch sizes of 32x32 pixels, and input image size of 224x224 pixels, a total of 49 patches used. We visually compare the results of our method to the second-best attribution method from our experiments - HiResCAM.

As anticipated, fidelity in attribution maps improves with a larger number of patches and decreases with fewer patches for both methods. However, our method exhibits significantly lower fidelity differences compared to HiResCAM. Even with less than one-tenth the patches, our method produces sparse and clear pixel-level attribution maps. This experiment highlights the advantage of using our attribution method in low-resource scenarios.

Finally, we demonstrate the versatility of our attribution method by applying it, without any modifications, to a BERT-based \cite{bert} text sentiment classification model. Figure \ref{fig:distilbert_contrastive} showcases the attribution visualizations for various movie and show reviews.

By employing these visualizations, we establish connections with our quantitative experiments and validate their robustness. The Local Consistency aspect of our metric assesses the method's ability to accurately identify salient areas in an image, expecting superior methods to primarily focus on pertinent regions corresponding to the model's training classes. Consequently, when examining Figure \ref{fig:vgg_contrastive}, we anticipate top-performing methods to concentrate on salient objects, minimizing attribution on background elements. Notably, LRP-$\epsilon$ and GradCAM exhibit a tendency to allocate a significant portion of their attribution outside the key objects of interest (e.g., dog and teddy bear in the first example, and cat, zucchini, and a grate in the second example). In contrast, methods such as absLRP, LRP-$\alpha_2\beta_1$, and GuidedGradCAM consistently maintain the bulk of their attribution on the relevant objects. This distinction is reflected in their average scores on the Local Consistency component of our metric, with LRP-$\epsilon$ and GradCAM acquiring comparatively lower scores (0.117 and 0.132) than absLRP, LRP-$\alpha_2\beta_1$, and GuidedGradCAM (0.396, 0.307, and 0.281).

To validate the Contrastiveness component of our metric, one can observe the differences in attribution maps when selecting different classes for visualization. This is reflected in the attribution maps' concentration on the selected target class, as opposed to dispersing attribution across all classes present in the image. Methods that display minimal alteration when altering the target class, such as LRP-$\epsilon$ and LRP-$\alpha_2\beta_1$, obtain lower Contrastiveness scores (0.022 and 0.016) compared to visually superior-performing methods, such as absLRP and cLRP (0.676 and 0.407).

Similar observations hold true for methods applied to other evaluated models, reinforcing the robustness and applicability of our proposed metric across diverse neural network architectures.

Our method stands out as the sole approach capable of consistently generating noise-free and contrastive attribution maps for each evaluated model, thereby providing readily interpretable visualizations.  In contrast, LRP-$\alpha_2 \beta_1$ and LRP-$\epsilon$ lack contrastiveness, attributing relevance indiscriminately. CAM-type methods, excluding GuidedGradCAM, struggle with isolating crucial image segments, attributing significance to background objects. Among top performers in VGG and ResNet50 experiments, excluding absLRP, GuidedGradCAM, and cLRP generate higher-quality sparse maps focusing on the target object. However, GuidedGradCAM introduces patchy noise, and cLRP consistently allocates relevance to background objects, possibly due to the model's susceptibility to relative magnitude attribution issues, as shown in Figure \ref{fig:intline_vs_alpha1beta0}.

The multitude of issues observed in existing methods, coupled with their notably reduced prominence, or complete absence, in absLRP, underscores the superior performance of our proposed method. Furthermore, the unparalleled versatility of absLRP, which can seamlessly adapt to diverse model architectures without requiring modifications, further emphasizes its superiority.

For the second experiment, we visually examine and analyze a subset of image examples from a randomly selected batch of 1024 images in the ImageNet dataset. Specifically, we focus on the VGG model and evaluate the ability of our evaluation metric to differentiate between high-quality and low-quality attribution maps. In this qualitative assessment, we present the findings through visualizations of the top scoring and bottom scoring examples.

\begin{figure}
  \centering
  \includegraphics[width=0.6\textwidth]{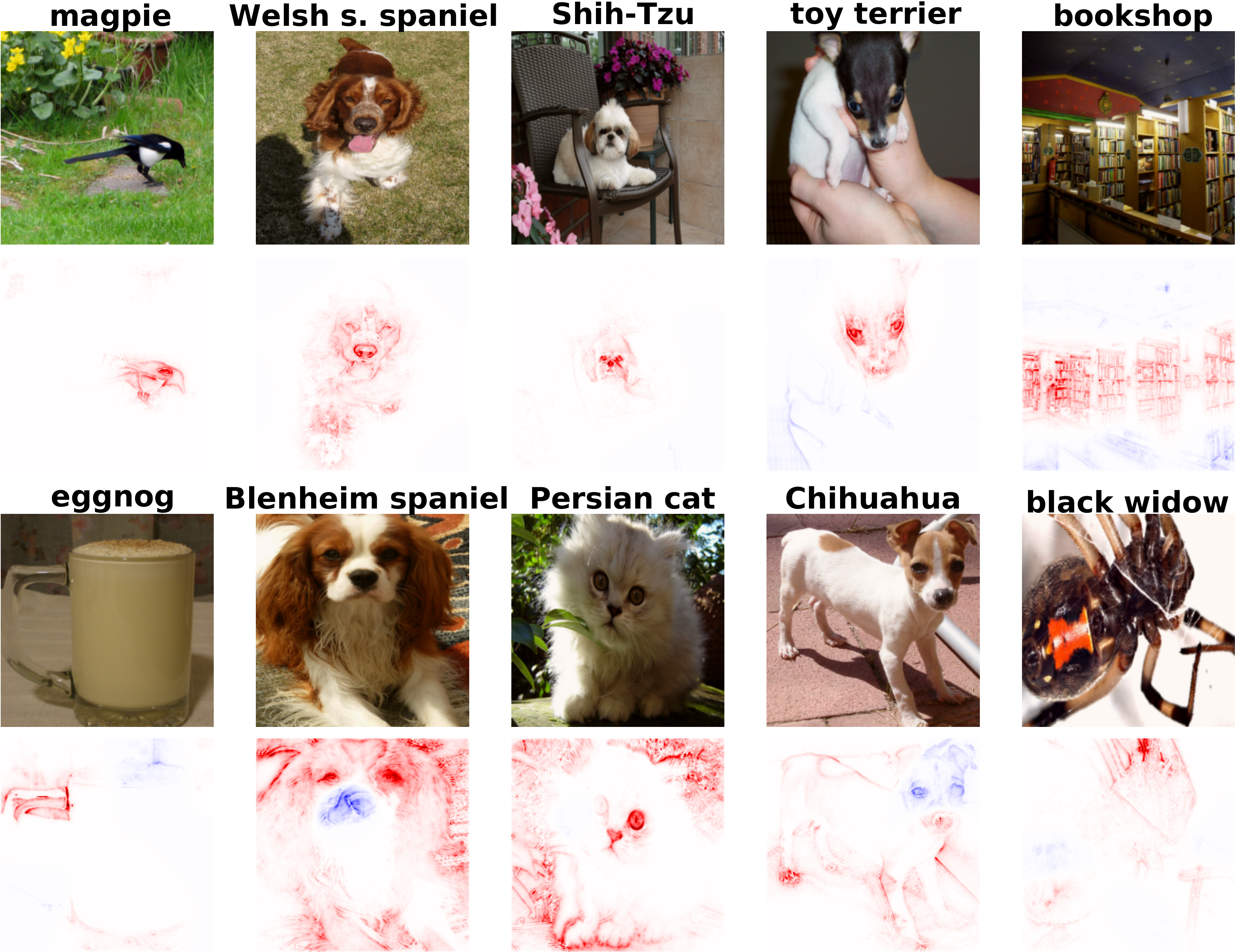}
  \Description{Four rows of images, every second row depicting an attribution map of the previous row. In the second row, the attributions are localized on the target object from the image. In the fourth row, the attributions are spread around the target object.}
  \captionof{figure}{Top five (top row) and bottom five (bottom row) scored attribution maps for our method in randomly sampled 1024 examples - ImageNet VGG.}
  \label{fig:vgg_top_bot}
\end{figure}


The results, depicted in Figure \ref{fig:vgg_top_bot}, demonstrate the effectiveness of our evaluation metric in distinguishing between high-quality and low-quality attribution maps. Top-performing attributions exhibit sparsity, absence of noise, and a clear focus on the target object, accurately capturing relevant features contributing to the model's decision. Conversely, the bottom performing attribution maps display less desirable traits. They exhibit a dispersed distribution of attribution values around the target objects and are plagued by noise. These attributes indicate a lack of precision and fidelity in capturing the salient features relevant to the model's decision-making process. Experiments for the ViT-Base architecture can be found in Appendix \ref{apx:vit_vis}, Figure \ref{fig:vit_top_bot}.

By visually assessing the top and bottom scoring examples, our experiment validates the effectiveness of our evaluation metric in distinguishing between high and low quality attribution maps. The results underline the importance of employing a robust and dependable evaluation metric to thoroughly assess the strengths and weaknesses of an attribution method.

\section{Conclusion}
In conclusion, this work makes two significant contributions. Firstly, we introduce a novel Layer-Wise Propagation rule, named Relative \textbf{Abs}olute Magnitude \textbf{L}ayer-Wise \textbf{R}elevance \textbf{P}ropagation (absLRP), that effectively addresses the issue of incorrect relative attribution between neurons within the same layer exhibiting varying absolute magnitude activations. We also apply the new rule without major modifications to three different architectures, including the recent Vision Transformer. Moreover, we propose the potential extension of this approach to diverse data types and tasks, such as text classification, opening up ample opportunities for future research and exploration.

Secondly, we propose a new evaluation method, \textbf{G}lobal \textbf{A}ttribution \textbf{E}valuation (GAE), which provides a fresh perspective on assessing the faithfulness, robustness and localization of attribution methods. In contrast to previous studies employing multiple metrics without a clear methodology for combining scores, our metric combines gradient-based masking and localization, providing a comprehensive evaluation using a single score. Through extensive experiments, we evaluate various attribution methods, revealing their individual strengths and weaknesses.

By quantitatively evaluating multiple attribution methods on diverse architectures and datasets, we establish the superiority of our approach over state-of-the-art and commonly used methods in this field. Furthermore, qualitative experiments conducted on both our attribution method and evaluation metric reinforce the advantages of our contributions.


\bibliographystyle{ACM-Reference-Format}
\bibliography{bibliography}


\appendix

\section{Additional visualizations for the VGG architecture}\label{apx:vgg_vis}

\begin{figure}[H]
  \centering
  \includegraphics[width=0.8\textwidth]{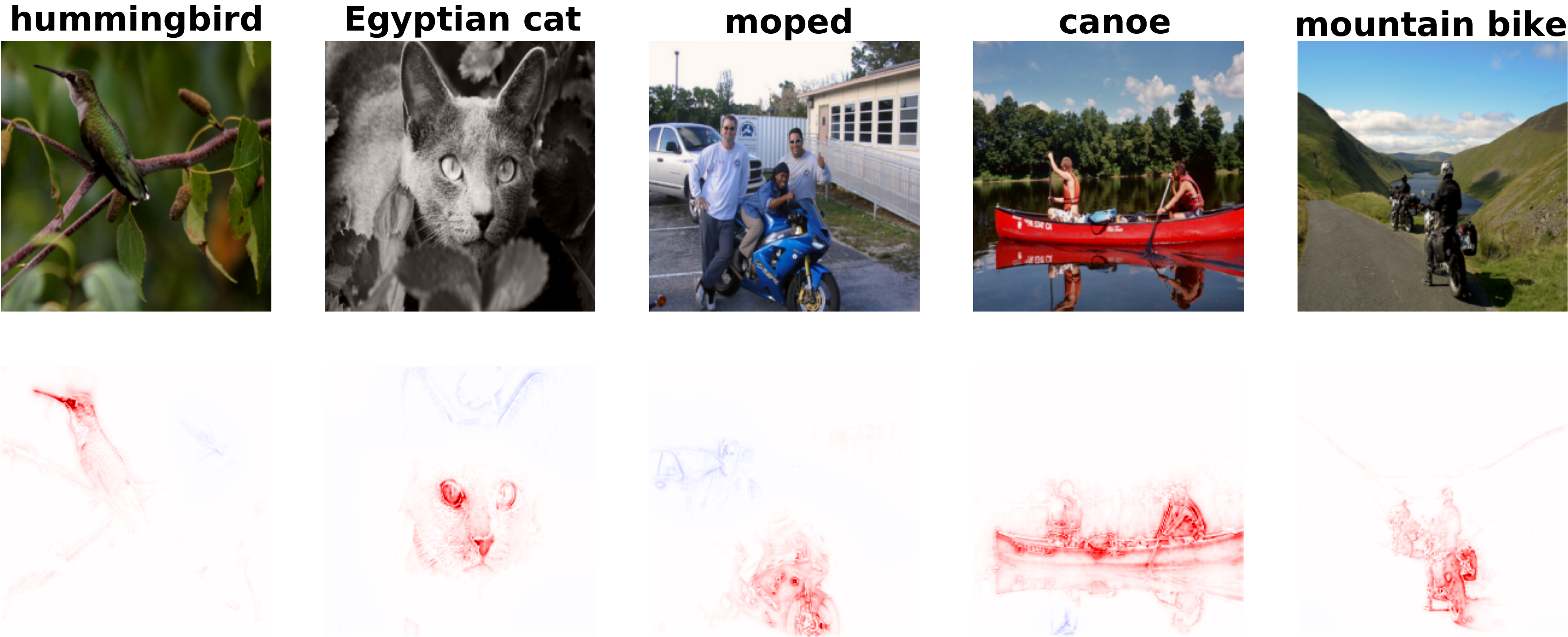}
  \Description{Two rows of five images. The first one depicts examples of different classes from the PascalVoc dataset. The second row depicts attribution maps for the aforementioned images. Attributions are localized on the target object.}
  \captionof{figure}{AbsLRP attribution maps for VGG - PascalVOC}
  \label{fig:vgg_simple_visualization_pascalvoc}
\end{figure}

\section{Additional visualizations for the Vision Transformer architecture}\label{apx:vit_vis}

\begin{figure}[H]
  \centering
  \includegraphics[width=0.8\textwidth]{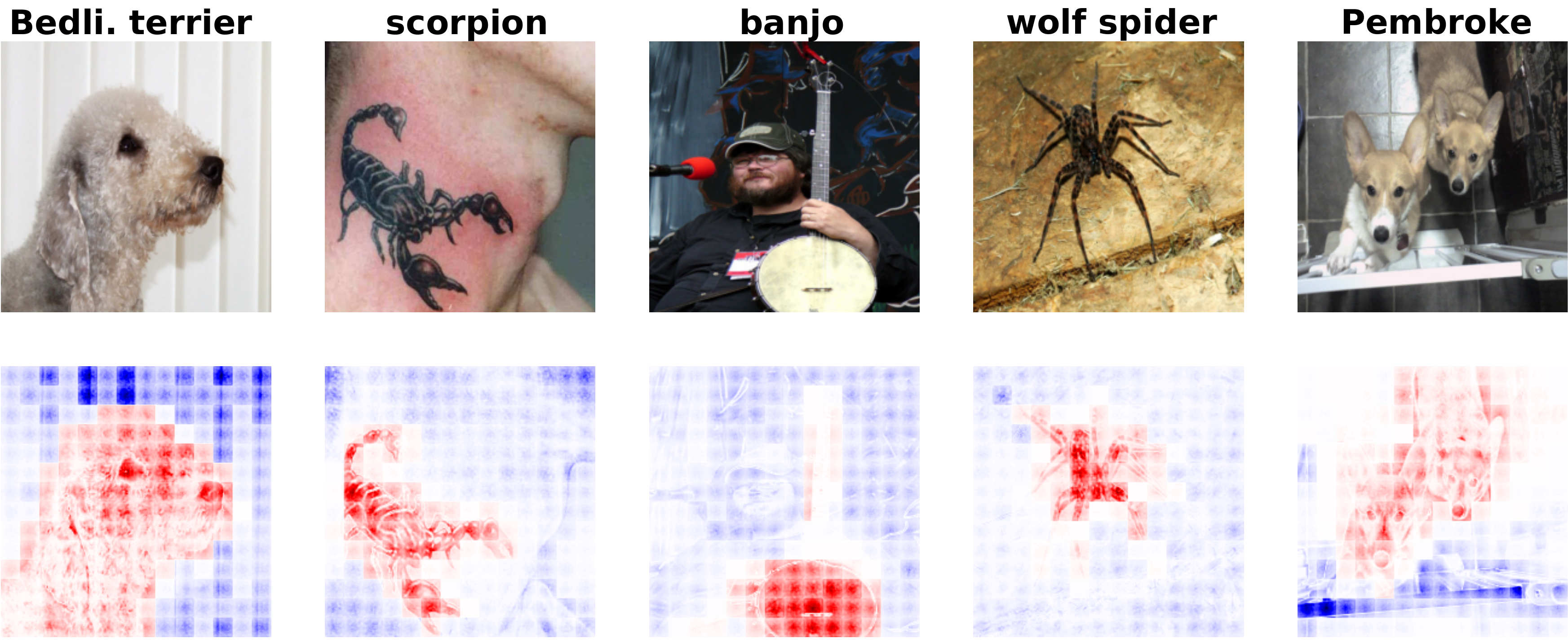}
  \Description{Two rows of five images. The first one depicts examples of different classes from the ImageNet dataset. The second row depicts attribution maps for the aforementioned images. Attributions are localized on the target object.}
  \captionof{figure}{AbsLRP attribution maps for ViT-Base - ImageNet}
  \label{fig:vit_simple_visualization_imagenet}
\end{figure}

\begin{figure}[H]
  \centering
  \includegraphics[width=0.8\textwidth]{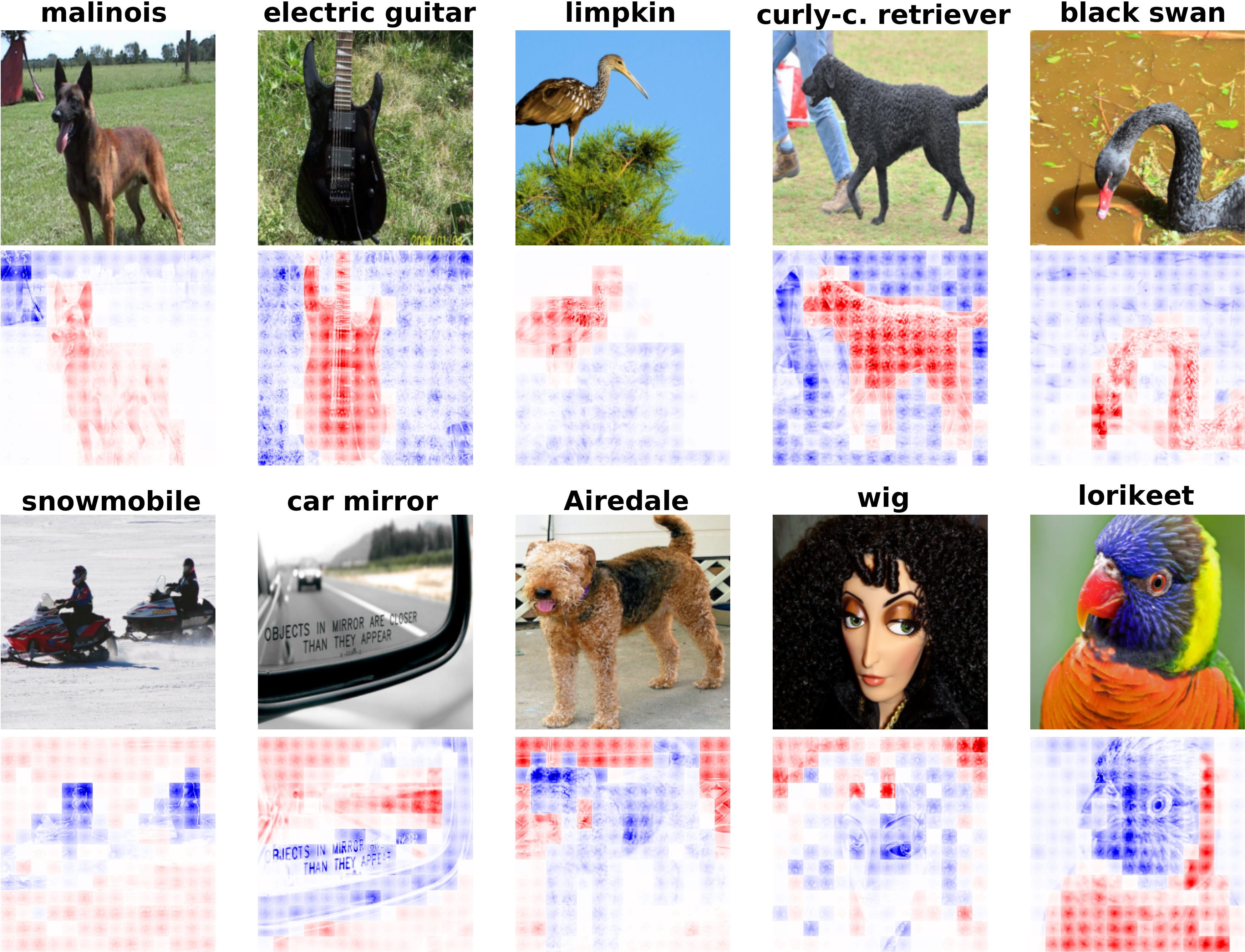}
  \Description{Four rows of images, every second row depicting an attribution map of the previous row. In the second row, the attributions are localized on the target object from the image. In the fourth row, the attributions are spread around the target object.}
  \captionof{figure}{Top five (top row) and bottom five (bottom row) scored attribution maps for our method in randomly sampled 1024 examples - ImageNet ViT-Base.}
  \label{fig:vit_top_bot}
\end{figure}

\section{Additional visualizations for the ResNet50 architecture}\label{apx:resnet_vis}

\begin{figure}[H]     
     \centering
    \includegraphics[width=0.8\textwidth]{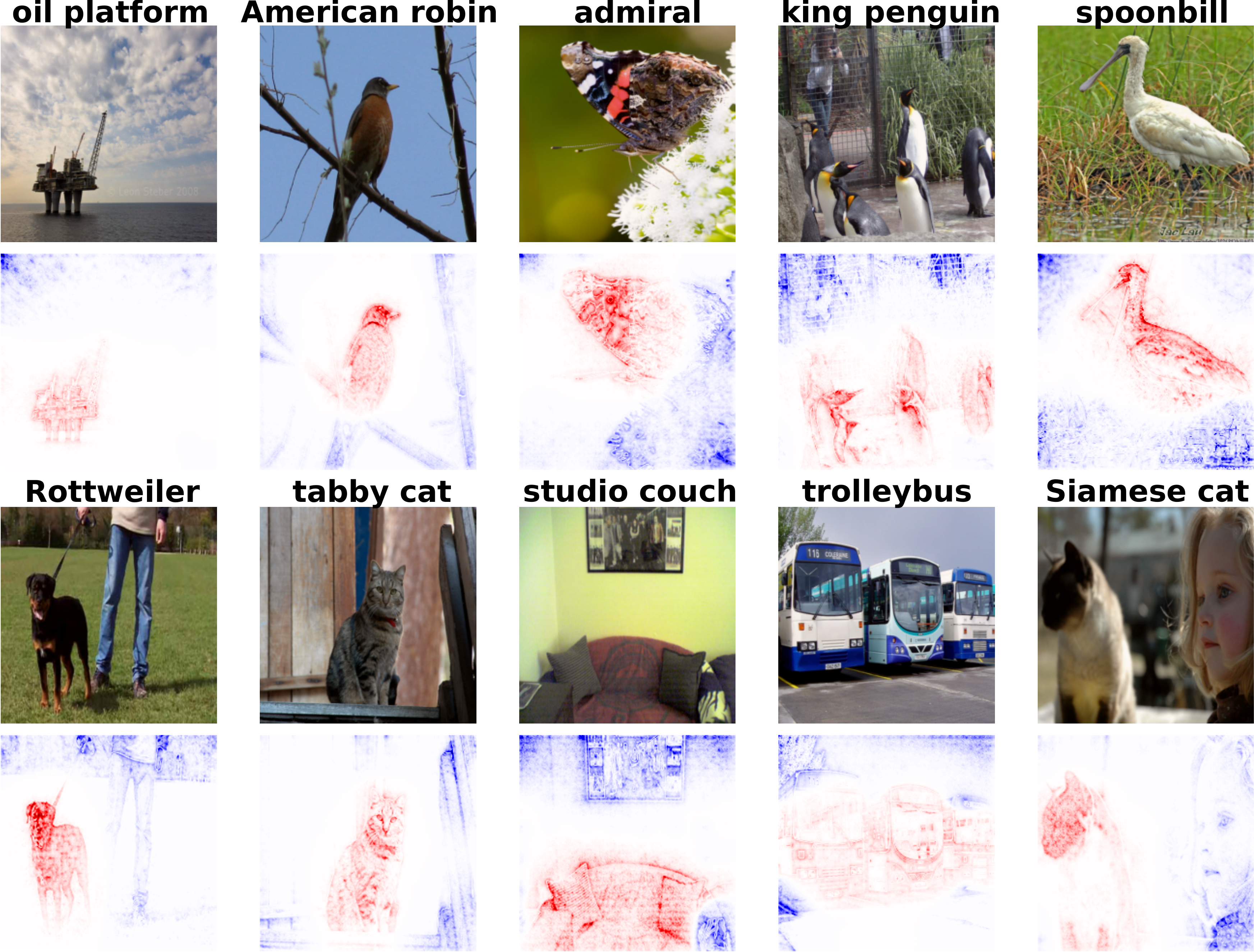}
     \Description{Four rows of five images, first and third row depicting examples of different classes from the two datasets. Second and fourth rows depicting attribution maps for the aforementioned images. Attributions are localized on the target object.}
     \captionof{figure}{AbsLRP attribution maps for ResNet50 - ImageNet (top), PascalVOC (bottom)}
     \label{fig:resnet50_simple_visualization_positive_negative}
\end{figure}

\begin{figure}[H]
\centering
     \includegraphics[width=0.8\textwidth]{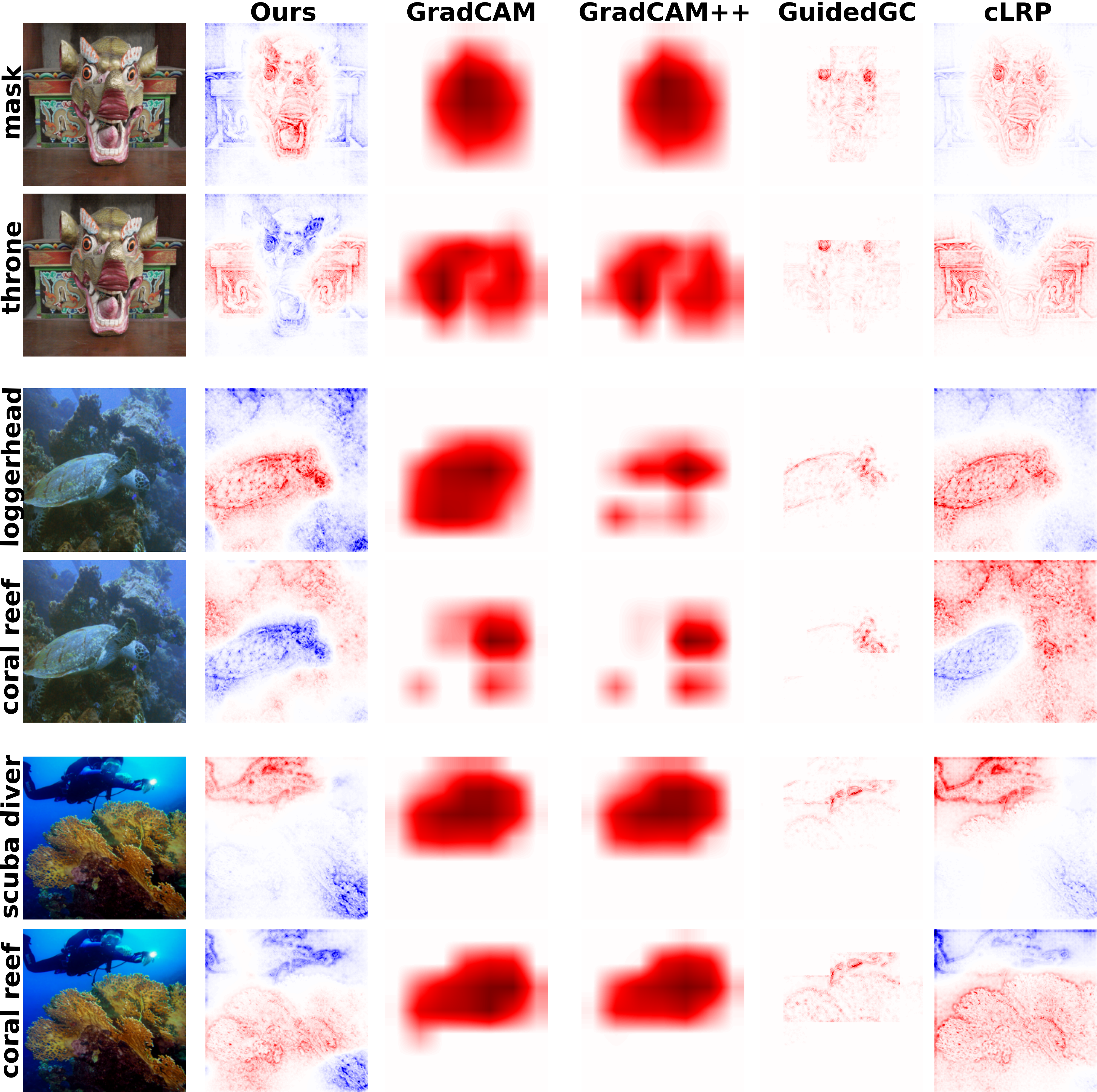}
     \Description{A grid of images and their respective attribution maps. Every pair of rows starts with the same image, but different target classes. In each row, the attributions of absLRP, GradCAM, GradCAM++, GuidedGradCAM and cLRP are visualized.}
     \captionof{figure}{Class-specific visualizations for multiple attribution methods - ImageNet ResNet50}
     \label{fig:resnet50_contrastive}
\end{figure}

\section{Additional results on commonly used metrics - PascalVOC}\label{apx:pascalvoc_apx}

\begin{table}[H]
    \caption{Scores obtained by various attribution methods on other metrics - VGG - PascalVOC dataset}
    \label{tab:vgg_results_other_pascalvoc}
    \begin{minipage}{\columnwidth}
    \begin{center}
    \begin{tabular}{c c c c c}
        \toprule
        Method name & $\overline{ROAD^{MoRF} (\downarrow)}$ & $\overline{ROAD^{LeRF} (\uparrow)}$ &$\overline{Lipschitz (\downarrow)}$ & $\overline{Focus (\uparrow)}$ \\ 
         \midrule
        Constant & 0.301 & 0.302 & \textbf{0.0} & 0.5 \\
        Random & 0.343 & 0.338 & 1.123 & 0.5 \\
        Saliency & 0.222 & 0.401 & 0.477 & 0.526 \\
        Input*Gradient & 0.22 & 0.361 & 0.394 & 0.537 \\
        Deconvolution & 0.252 & 0.416 & 0.47 & 0.5 \\
        LRP-$\epsilon$ & 0.166 & 0.428 & 0.361 & 0.537 \\
        DeepLIFT & 0.108 & 0.478 & 0.316 & 0.594 \\
        LRP-$\alpha_1\beta_0$ & 0.105 & 0.562 & 0.124 & 0.507 \\
        LRP-$\alpha_2\beta_1$ & 0.105 & \textbf{0.58}* & 0.089 & 0.526 \\
        Integrated Gradients & 0.182 & 0.419 & 0.361 & 0.549 \\
        SmoothGrad & 0.133 & 0.533 & 0.375 & 0.528 \\
        GradCAM & \textbf{0.074}* & 0.511 & 0.775 & 0.766 \\
        HiResCAM & 0.102 & 0.52 & 0.73 & 0.731 \\
        LayerCAM & \textbf{0.087}* & 0.53 & 0.613 & 0.581 \\
        GradCAM++ & \textbf{0.084}* & 0.51 & 0.6 & 0.572 \\
        GuidedGradCAM & \textbf{0.071}* & \textbf{0.591}* & 0.135 & \textbf{0.779}* \\
        cLRP & \textbf{0.062}* & \textbf{0.559}* & 0.17 & 0.755 \\
        RAP & 0.103 & 0.481 & 0.195 & 0.441 \\
        cRAP & 0.134 & 0.468 & 0.346 & 0.761 \\
        \textbf{Ours} & \textbf{0.071}* & \textbf{0.563}* & 0.237 & \textbf{0.787}* \\
         \bottomrule
    \end{tabular}
\end{center}
\end{minipage}
\footnotesize $\downarrow$ - lower score is better; $\uparrow$ - higher score is better \\
\footnotesize *Difference in results could not be confirmed by Wilcoxon signed-rank test at p=0.05
\end{table}

\begin{table}[H]
    \caption{Scores obtained by various attribution methods on other metrics - ResNet50 - PascalVOC dataset}
    \label{tab:resnet50_results_other_pascalvoc}
    \begin{minipage}{\columnwidth}
    \begin{center}
    \begin{tabular}{c c c c c}
        \toprule
        Method name & $\overline{ROAD^{MoRF} (\downarrow)}$ & $\overline{ROAD^{LeRF} (\uparrow)}$ &$\overline{Lipschitz (\downarrow)}$ & $\overline{Focus (\uparrow)}$ \\ 
         \midrule
        Integrated Gradients & 0.248 & 0.548 & 0.339 & 0.639 \\
        SmoothGrad & 0.187 & \textbf{0.655}* & 0.335 & 0.633 \\
        GradCAM & 0.152 & 0.585 & 0.472 & 0.735 \\
        HiResCAM & 0.153 & 0.567 & 0.471 & 0.67 \\
        LayerCAM & 0.154 & 0.584 & 0.495 & 0.66 \\
        GradCAM++ & 0.146 & 0.584 & 0.505 & 0.733 \\
        GuidedGradCAM & 0.141 & 0.63 & \textbf{0.131}* & 0.743 \\
        cLRP & \textbf{0.109}* & \textbf{0.683}* & \textbf{0.142}* & \textbf{0.906}* \\
        RAP & 0.218 & 0.413 & 0.796 & 0.524 \\
        cRAP & 0.173 & 0.566 & 0.693 & 0.784 \\
        \textbf{Ours} & \textbf{0.113}* & \textbf{0.659}* & 0.232 & \textbf{0.913}* \\
         \bottomrule
    \end{tabular}
\end{center}
\end{minipage}
\footnotesize $\downarrow$ - lower score is better; $\uparrow$ - higher score is better \\
\footnotesize *Difference in results could not be confirmed by Wilcoxon signed-rank test at p=0.05
\end{table}

\begin{table}[H]
    \caption{Scores obtained by various attribution methods on other metrics - ViT-Base patch size 16 - PascalVOC dataset}
    \label{tab:vitbase_results_other_pascalvoc}
    \begin{minipage}{\columnwidth}
    \begin{center}
    \begin{tabular}{c c c c c}
        \toprule
        Method name & $\overline{ROAD^{MoRF} (\downarrow)}$ & $\overline{ROAD^{LeRF} (\uparrow)}$ &$\overline{Lipschitz (\downarrow)}$ & $\overline{Focus (\uparrow)}$ \\ 
         \midrule
        GradCAM & \textbf{0.131} & \textbf{0.642}* & 0.853 & \textbf{0.8}* \\
        GradCAM++ & 0.341 & 0.451 & 2.434 & 0.594 \\
        HiResCAM & 0.162 & \textbf{0.634}* & 0.524 & \textbf{0.785}* \\
        LayerCAM & 0.182 & \textbf{0.62}* & 0.586 & 0.756 \\
        LLAttention & 0.208 & 0.566 & 0.574 & 0.485 \\
        Rollout & 0.232 & 0.572 & 1.059 & 0.487 \\
        TIBAV & 0.168 & \textbf{0.609}* & 0.756 & 0.677 \\
        \textbf{Ours} & 0.195 & 0.573 & \textbf{0.46} & 0.689 \\
         \bottomrule
    \end{tabular}
\end{center}
\end{minipage}
\footnotesize $\downarrow$ - lower score is better; $\uparrow$ - higher score is better \\
\footnotesize *Difference in results could not be confirmed by Wilcoxon signed-rank test at p=0.05
\end{table}

\end{document}